\documentclass[lettersize,journal]{IEEEtran}
\usepackage{amsmath,amsfonts}
\usepackage{algorithmic}
\usepackage{algorithm}
\usepackage{array}
\usepackage[caption=false,font=normalsize,labelfont=sf,textfont=sf]{subfig}
\usepackage{textcomp}
\usepackage{stfloats}
\usepackage{url}
\usepackage{verbatim}
\usepackage{graphicx}
\usepackage[nocompress]{cite}
\usepackage[hidelinks]{hyperref}
\usepackage{siunitx}
\usepackage{multirow}
\usepackage{array}
\begin{document}

\title{Underactuated Robotic Hand with Grasp State Estimation \\ Using Tendon-Based Proprioception}

\author{Jae-Hyun Lee,~\IEEEmembership{Student Member,~IEEE},
        Jonghoo Park,
        and Kyu-Jin Cho,~\IEEEmembership{Member,~IEEE}
\thanks{This work was supported by the National Research Foundation of Korea(NRF) Grant funded by the Korean Government(MSIT) (RS-2023-00208052).}  
\thanks{Jae-Hyun Lee and Kyu-Jin Cho are with the BioRobotics Laboratory (BRL), Soft Robotics Research Center (SRRC), Institute of Advanced Machines and Design (IAMD), Department of Mechanical Engineering, Institute of Engineering, Seoul National University (SNU), Seoul 08826, Korea (e-mail: lajaha37@snu.ac.kr; kjcho@snu.ac.kr).}
\thanks{Jonghoo Park is with the Robotics Development Team, Hyundai Mobis, Seoul 06232, Korea (e-mail: jonghoo.park@mobis.com).}
}



\maketitle

\begin{abstract}
Anthropomorphic underactuated hands are valued for their structural simplicity and inherent adaptability. However, the uncertainty arising from interdependent joint motions makes it challenging to capture various grasp states during hand--object interaction without increasing structural complexity through multiple embedded sensors. This motivates the need for an approach that can extract rich grasp-state information from a single sensing source while preserving the simplicity of underactuation. This study proposes an anthropomorphic underactuated hand that achieves comprehensive grasp state estimation, using only tendon-based proprioception provided by series elastic actuators (SEAs). Our approach is enabled by the design of a compact SEA with high accuracy and reliability that can be seamlessly integrated into sensorless fingers. By coupling accurate proprioceptive measurements with potential energy-based modeling, the system estimates multiple key grasp state variables, including contact timing, joint angles, relative object stiffness, and external disturbances. Finger-level experimental validations and extensive hand-level grasp functionality demonstrations confirmed the effectiveness of the proposed approach. These results highlight tendon-based proprioception as a compact and robust sensing modality for practical manipulation without reliance on vision or tactile feedback.
\end{abstract}


\begin{figure*}[t]
\centering
\includegraphics[width=0.87\textwidth]{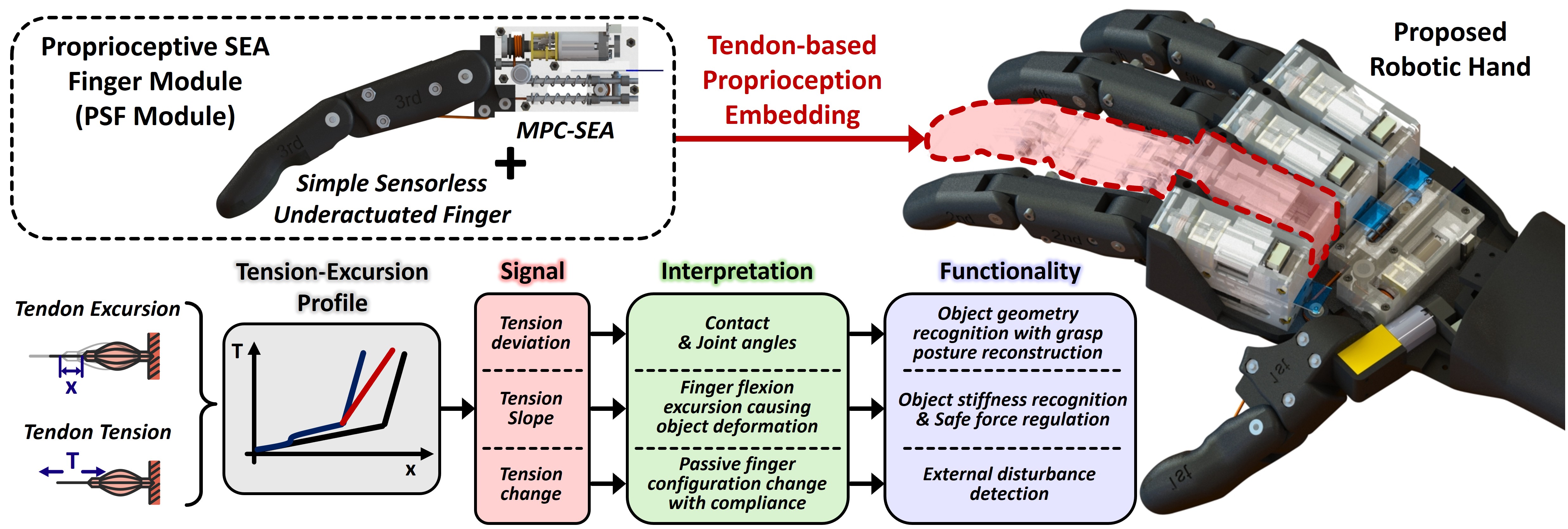}
\caption{Overview of the proposed anthropomorphic underactuated hand with tendon-based proprioception embedding through PSF modules.}
\label{fig_1}
\end{figure*}

\section{Introduction}
\IEEEPARstart{A}{nthropomorphic} robotic hands have been widely adopted to replicate the functionality of the human hand.
Among various actuation strategies, underactuated hands are extensively employed due to their structural simplicity and adaptability to diverse object geometries~\cite{birglen_underactuated_2007, piazza_century_2019}.
As manipulation fundamentally involves physical interaction between the hand and the object, it is essential to sense grasp states throughout this interaction~\cite{fan_understanding_2022}.
For instance, capturing finger configuration and force information is crucial for understanding object properties and interaction dynamics, thereby supporting the decision-making process during manipulation~\cite{homberg_robust_2019}.
In underactuated hands, however, the inherent uncertainty arising from interdependent joint motions complicates grasp state sensing~\cite{birglen_underactuated_2007, belzile_stiffness_2016}.
For this reason, key grasp state variables, such as object contact, joint angles, and internal forces, are difficult to capture without embedding multiple dedicated sensors within each finger.

Yet, embedding multiple sensors into each finger makes the system more complex and less efficient, thereby undermining the fundamental advantage of underactuated mechanisms, which lies in their simple structure.
This motivates sensing approaches that exploit a single comprehensive sensing signal, providing an effective alternative tailored to the characteristics of underactuation.
In this context, proprioception, which provides internal information about force or position in the human muscle-tendon system~\cite{tuthill_proprioception_2018}, is a compelling sensing modality for underactuated hands over other modalities, such as vision and touch.
Vision does not provide direct force feedback, making it difficult to understand hand--object interaction, and it is also vulnerable to occlusion once the hand closes around the object~\cite{calandra_more_2018, boruah_shape_2023}.
Tactile sensors, which require additional mounting and wiring on the fingers, increase structural complexity and offer only local information rather than comprehensive information across the entire finger~\cite{kim_multiparameter_2024, yi_underactuated_2023}.

Therefore, proprioception has been widely employed in various robotic hands and grippers as a practical sensing alternative.
In previous work, motor current has been used as a surrogate for proprioception, but its accuracy is constrained by electrical noise and transmission friction, which hinder reliable force sensing~\cite{hu_force_2024, yeh_compliant_2024}. 
More direct approaches--such as load cells~\cite {yan_texture_2022} or torque sensors~\cite{belzile_stiffness_2016, arolovitch_kinesthetic-based_2024}, fiber Bragg gratings (FBG) sensors~\cite{yi_underactuated_2023}, and series elastic actuators (SEAs)~\cite{yan_sea-based_2023, kaya_series_2022, hua_design_2023, chen_proprioception-based_2018, wu_back-drivable_2023}--have been explored for measuring internal actuation force.
Among them, several studies have applied proprioceptive sensing to underactuated hands through different approaches.
Some reported force profiles associated with object contact or stiffness, focusing primarily on illustrating overall trends of how force signals change during interaction~\cite{yi_underactuated_2023, hua_design_2023, wu_back-drivable_2023}.
Others relied solely on data-driven methods to infer object-related properties from proprioceptive signals, without incorporation of physical interpretation of underactuation~\cite{yan_texture_2022, arolovitch_kinesthetic-based_2024}.
There also exist modeling-based approaches that investigated underactuated finger motion with joint stiffness to estimate contact location under specific hardware conditions~\cite{belzile_stiffness_2016}.

However, previous studies offered only limited interpretations of an implicit proprioceptive signal that aggregates information from multiple joints.
Consequently, none of these works presented a generalized estimation method that leverages the characteristics of underactuation to extract diverse grasp states from a single sensing signal.
Moreover, many existing implementations result in form factors of bulky grippers or hands with enlarged bases and palms, severely limiting their applicability in anthropomorphic hands.
Therefore, achieving compact in-hand integration of proprioceptive sensing while ensuring high sensing performance also remains challenging.

To address the aforementioned challenges, we present an anthropomorphic underactuated robotic hand that achieves comprehensive grasp state estimation during hand--object interaction and realizes practical grasp functionalities utilizing only tendon-based proprioception.
To this end, we designed a miniature proprioceptive cable-driven SEA (MPC-SEA) that can be seamlessly integrated within the hand.
Its compactness is achieved through a simplified sensing and guiding scheme relative to conventional SEA structures, while ensuring tendon tension sensing with high accuracy and reliability.
When attached to a simple underactuated finger without any tactile sensors or joint encoders, the MPC-SEA forms a proprioceptive SEA finger (PSF) module that enables compact in-hand integration while providing immediate sensing capability to otherwise sensorless hands, as shown in Fig.~\ref{fig_1}.

The PSF module acquires real-time tendon excursion, which refers to the amount of tendon pulled by the actuator, and tendon tension. 
These tendon-based proprioceptive signals are combined with potential energy--based modeling to estimate physically interpretable grasp state variables from the implicit tension profile of an underactuated finger.
From a single tension--excursion profile, the hand can detect the timing of proximal and distal contacts, estimate each joint angle, extract relative stiffness information, and identify external disturbances through SEA-driven compliance.
All functions are executed in real time through a lightweight rule-based algorithm that combines computational simplicity with effective state estimation and stable grasp control.
The proposed approach was validated through finger-level experiments---confirming finger modeling results and tension profile characterization---as well as extensive hand-level demonstrations, including grasp posture reconstruction of irregular objects, safe grasp force regulation for soft deformable objects without fracture, and blind grasping with proprioceptive-only recognition of objects varying in size, shape, and stiffness.

\section{Hardware Design}
\subsection{MPC-SEA Design}

\begin{figure}[t]
\centering
\includegraphics[width=0.85\columnwidth]{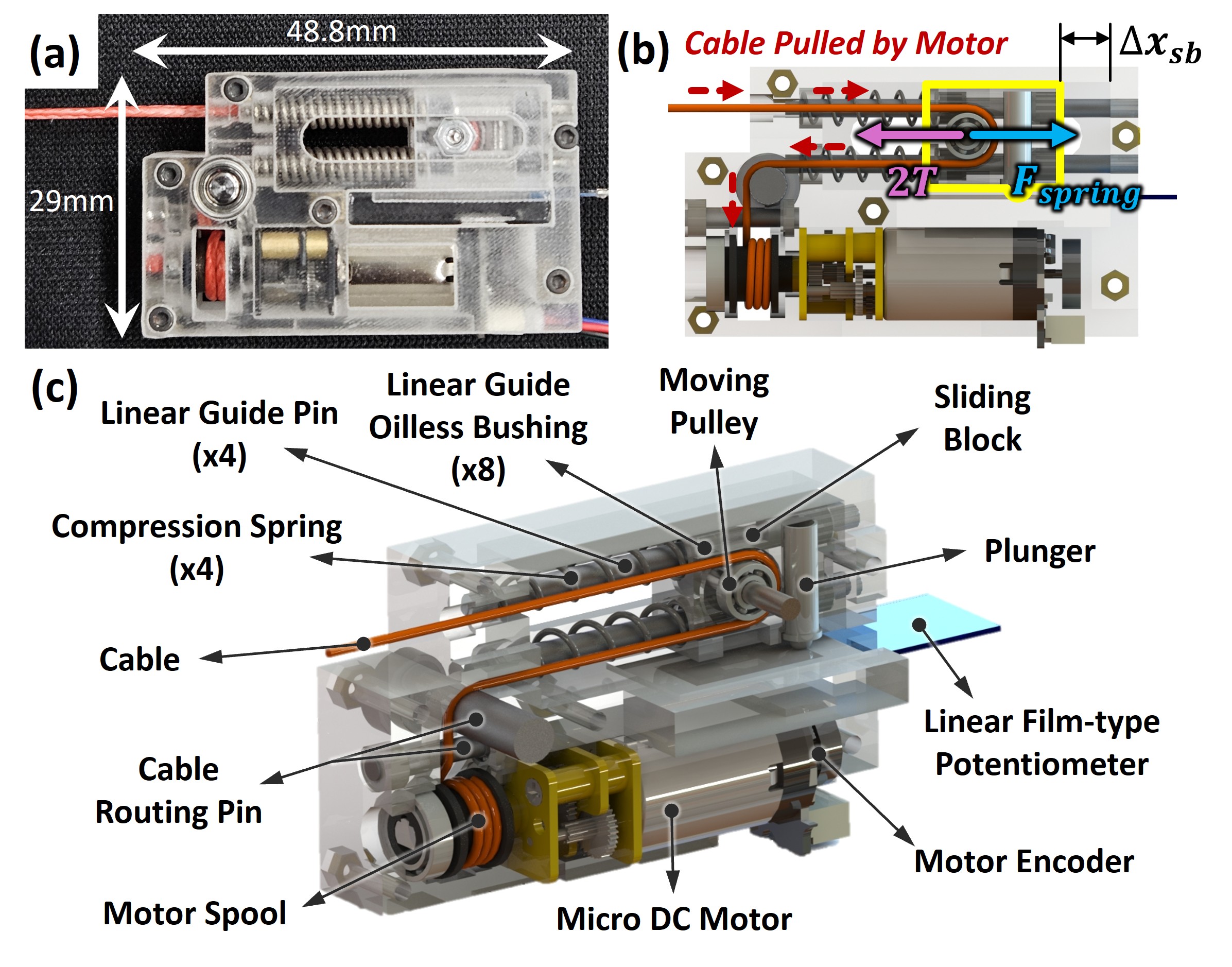}
\caption{Design of MPC-SEA. (a) Appearance and key specifications. (b) Structure and components. (c) Actuation mechanism.}
\label{fig_2}
\end{figure}

The physical appearance of the MPC-SEA is shown in Fig.~\ref{fig_2}(a). The module measures \SI{48.8}{\milli\metre} $\times$ \SI{29.0}{\milli\metre} $\times$ \SI{17.6}{\milli\metre} and weighs \SI{37.1}{\gram}, making it highly compact and lightweight. It can deliver up to \SI{49}{\newton} of cable tension and pull the cable at a speed of \SI{28.9}{\milli\metre\per\second}.
The structure of the MPC-SEA is shown in Fig.~\ref{fig_2}(c). Actuation is achieved by winding a cable (Dynext PU \SI{0.9}{\milli\metre}, Amare Ropes, Italy) around a motor spool driven by a DC motor with a magnetic encoder (380:1 Micro Metal Gearmotor with 12 CPR Encoder, Pololu, USA). The cable is routed through a moving pulley housed in a sliding block, which translates linearly and compresses the spring when tension is applied, as shown in Fig.~\ref{fig_2}(b). Oilless bushings and pins guide linear motion, and the displacement of the sliding block is measured using a film-type potentiometer (ThinPot, Spectra Symbol, USA) via a spring-loaded plunger, ensuring consistent surface contact. The measured displacement is converted into cable tension using Hooke’s law, with the effective spring constant determined by four parallel springs and the two-to-one pulley mechanism, as expressed in \eqref{eq:sea1}, \eqref{eq:sea2}, \eqref{eq:sea3}.



\begin{align}
\Delta x_{\mathrm{sb}} &= \frac{V_{\mathrm{o}}}{V_{\mathrm{ref}}} L_{p} - x_{0} \label{eq:sea1} \\
F_{\mathrm{spring}} &= 4 \times k_{\mathrm{sea}} \, \Delta x_{\mathrm{sb}} \label{eq:sea2} \\
T &= \frac{F_{\mathrm{spring}}}{2} = \frac{2 k_{\mathrm{sea}} L_{p}}{V_{\mathrm{ref}}} V_{\mathrm{o}} - 2 k_{\mathrm{sea}} x_{0} \label{eq:sea3}
\end{align}

where $\Delta x_{\mathrm{sb}}$ is the displacement of the sliding block, $F_{\mathrm{spring}}$ is the restoring force of springs, and $T$ is the cable tension. $V_{\mathrm{ref}}$ is the voltage reference, $V_{\mathrm{o}}$ is the voltage output, $L_{p}$ is the potentiometer length, $x_{0}$ is the initial position of the sliding block, and $k_{\mathrm{sea}}$ is the spring constant of a single spring.

The linear motion is implemented using oilless bushings and guide pins instead of a conventional ball screw, and displacement sensing relies on a film-type potentiometer with a plunger. This design selection enables significant miniaturization compared to conventional linear SEA designs, while maintaining high sensing accuracy and reliability.
As a self-contained module, the MPC-SEA can be used immediately by simply connecting the cable to a finger.

\subsection{Finger and Thumb Design}

\begin{figure}[t]
\centering
\includegraphics[width=0.8\columnwidth]{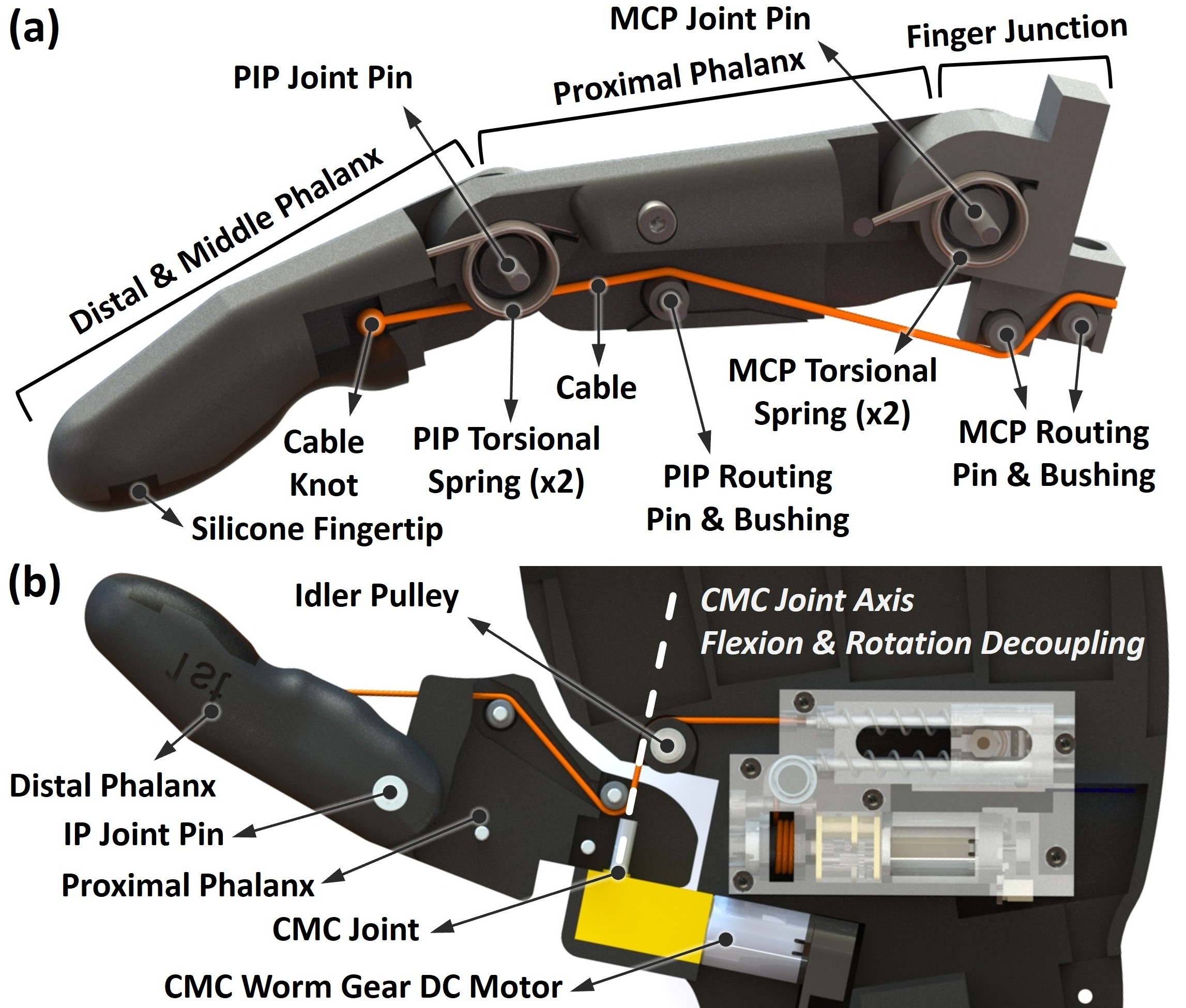}
\caption{Design of the finger and the thumb. (a) Finger (2nd--5th). (b) Thumb.}
\label{fig_3}
\end{figure}

All fingers (2nd--5th), except for the thumb (1st), share an identical design with only slight differences in dimensional parameters. The structure and components of the sensorless finger are shown in Fig.~\ref{fig_3}(a). Each finger consists of two joint degrees of freedom---metacarpophalangeal (MCP) and proximal interphalangeal (PIP) joints. Finger flexion is underactuated with tendon routing guided by oilless bushings, while extension is passively achieved via two torsional springs per joint. The fingertip surface is made of silicone (KE-1300T, Shin-Etsu Chemical, Japan) to increase friction.

Unlike the other fingers, the thumb features a single degree-of-freedom interphalangeal (IP) joint for flexion and an additional carpometacarpal (CMC) rotation joint for opposition, as shown in Fig.~\ref{fig_3}(b). The CMC rotation is actuated by a worm-gear DC motor (603:1 WG12F-1510E, Motorbank, Korea). Due to spatial constraints, the thumb cannot be directly connected to the MPC-SEA; instead, it is routed through an idler pulley. In this configuration, the tendon path passes through the CMC joint axis, thereby decoupling flexion from rotation.
The specifications of the SEA and the hand are summarized in Table~\ref{tab:spec_table}.

\begin{table}[t]
\centering
\caption{Specifications of the Hardware}
\label{tab:spec_table}
\renewcommand{\arraystretch}{1.15}
\begin{tabular}{c|c|c}
\hline
\textbf{Category} & \textbf{Feature} & \textbf{Value} \\ \hline

\multirow{3}{*}{Size}
 & SEA Dimension [\SI{}{\milli\metre}] & 48.8 × 29.0 × 17.6 \\
 & Finger Length (3rd) [\SI{}{\milli\metre}] & 98.5 \\
 & Hand Length [\SI{}{\milli\metre}] & 213.5 \\ \hline

\multirow{4}{*}{Weight}
 & SEA Weight [\SI{}{\gram}] & 37.1 \\
 & Finger Weight [\SI{}{\gram}] & 28.6 \\
 & Hand Weight [\SI{}{\gram}] & 513 \\ \hline

\multirow{5}{*}{Actuation}
 & SEA Max. Tension [\SI{}{\newton}] & 49 \\
 & SEA No-Load Speed [\SI{}{\milli\metre\per\second}] & 28.9 \\
 & Finger Closing Time [\SI{}{\second}] & 1.3 \\
 & SEA Spring Constant [\SI{}{\newton\per\milli\metre}] & 5.44 \\
 \hline

\multirow{8}{*}{Mechanism}
 & Joint Degrees of Freedom & 10 \\
 & Number of Actuators & 6 \\
 & Finger MCP / PIP Joint RoM [\SI{}{\degree}] & 90 / 100 \\
 & Finger Joint Initial Angle [\SI{}{\degree}] & 15 \\
 & Thumb IP Joint RoM [\SI{}{\degree}] & 77 \\
 & Thumb IP Joint Initial Angle [\SI{}{\degree}] & 10 \\
 & Finger Joint Stiffness [\SI{}{\newton\milli\metre\per\degree}] & 0.86 \\ \hline

\end{tabular}
\end{table}

\subsection{Electronics Integration}

\begin{figure}[t]
\centering
\includegraphics[width=0.8\columnwidth]{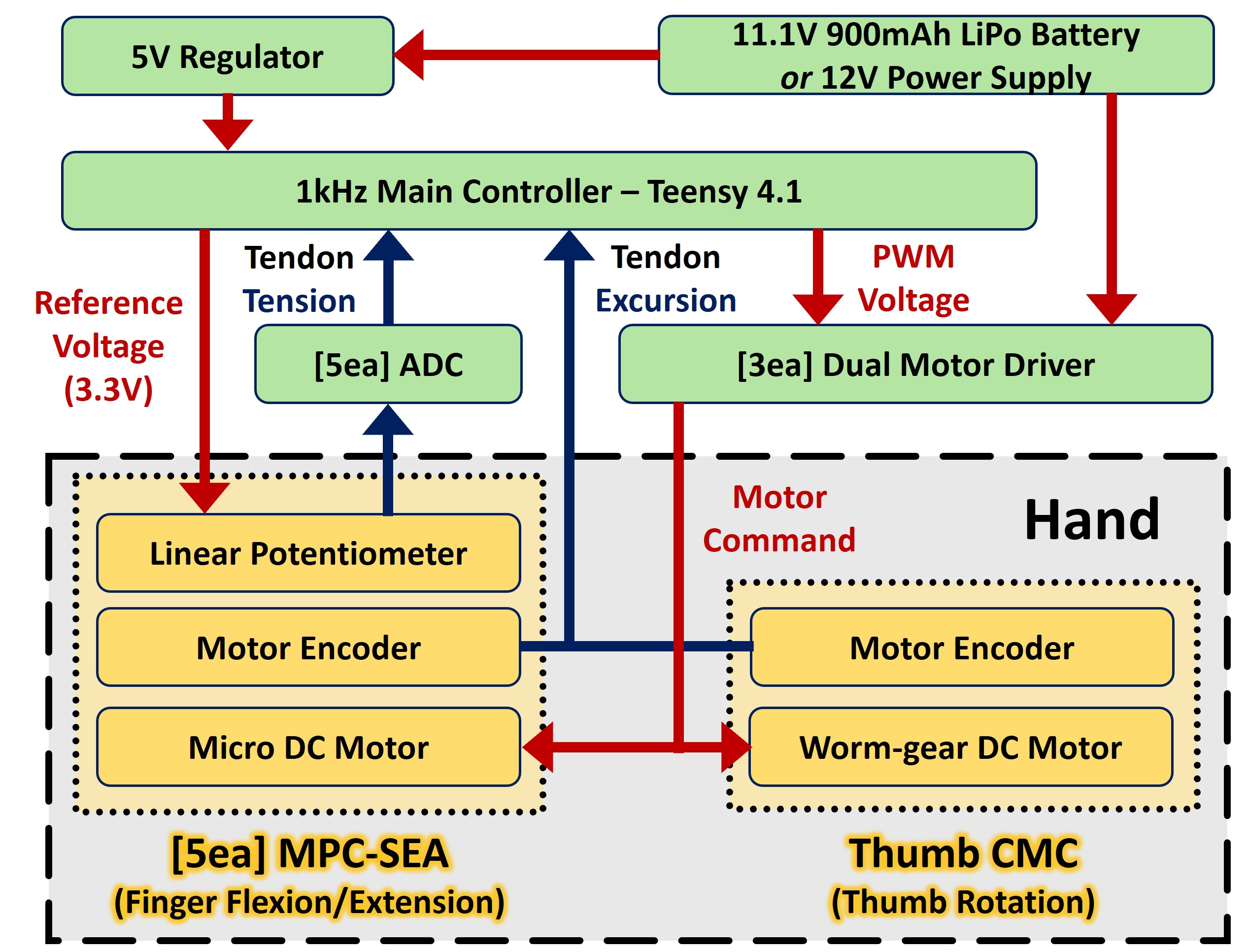}
\caption{Electronic system architecture of the proposed hand.}
\label{fig_4}
\end{figure}

Fig.~\ref{fig_4} illustrates the overall electronic system architecture of the proposed hand.
The system is powered either by an \SI{11.1}{\volt} Li-Po battery (\SI{900}{mAh}) or a \SI{12}{\volt} external power supply, which is regulated to \SI{5}{\volt} through a D24V10F5 regulator for all low-voltage electronics.
 A Teensy 4.1 serves as the main controller, executing the \SI{1}{\kilo\hertz} sensing and motor-control loops. 
The controller provides the reference voltage $V_{\mathrm{ref}}$ of \SI{3.3}{\volt} and receives tendon tensions from potentiometers via ADS1015 12-bit ADCs.
Tendon excursion is obtained directly from the motor encoder signals embedded in the MPC-SEAs.
For actuation, TB6612FNG motor drivers apply PWM-driven motor currents.
This architecture enables synchronized multi-finger actuation with real-time proprioceptive feedback from every SEA module.

\section{Modeling and Characterization}

\subsection{Finger Modeling}
To model the relationship between tendon excursion, tendon tension, and joint angles of the two underactuated joints, the minimum potential energy principle was applied, consistent with previous approaches in~\cite{zhang_dexterity_2020, odhner_compliant_2014}.
Friction was neglected, and quasi-static movement was assumed.
Table~\ref{tab:nomenclature} lists the symbols used in the finger modeling process.
The system’s total energy $E_{\mathrm{Total}}$ comprises the elastic energy $E_{\mathrm{e}}$ from the SEA spring and the extension torsional springs, along with gravitational energy $E_{\mathrm{g}}$, as expressed in \eqref{eq:energy_elastic}, \eqref{eq:energy_gravity}, \eqref{eq:energy_total}.

\begin{align}
E_{\mathrm{e}}(\theta_1, \theta_2, x)
&= 4 \cdot 0.5 \cdot k_{\mathrm{sea}} \cdot
\left[ \frac{x - \ell(\theta_1, \theta_2)}{2} \right]^{2} \notag\\
& + 2 \cdot 0.5 \cdot k_1 \cdot \left( \theta_{\mathrm{pre1}} + \theta_1 \right)^{2} \notag\\
& + 2 \cdot 0.5 \cdot k_2 \cdot \left( \theta_{\mathrm{pre2}} + \theta_2 \right)^{2}
\label{eq:energy_elastic} \\[4pt]
E_{\mathrm{g}}(\theta_1, \theta_2)
&= m_1 \cdot g \cdot
\left( \mathbf{\hat{z}}^\top R_{\mathrm{hand}} \, R_{\mathrm{j}}(\theta_{\mathrm{i1}} + \theta_1) \, L_{\mathrm{c1}} \right) \notag\\
& + m_2 \cdot g \cdot
\left[ \mathbf{\hat{z}}^\top R_{\mathrm{hand}} \,
\big( R_{\mathrm{j}}(\theta_{\mathrm{i1}} + \theta_1) \, L_1 \right. \notag\\
& \left. + \, R_{\mathrm{j}}(\theta_{\mathrm{i1}} + \theta_1 + \theta_{\mathrm{i2}} + \theta_2) \, L_{\mathrm{c2}} \big) \right]
\label{eq:energy_gravity} \\[4pt]
E_{\mathrm{Total}}(\theta_1, \theta_2, x)
&= E_{\mathrm{e}}(\theta_1, \theta_2, x)
+ E_{\mathrm{g}}(\theta_1, \theta_2)
\label{eq:energy_total}
\end{align}


\begin{figure*}[t]
\centering
\includegraphics[width=0.9\textwidth]{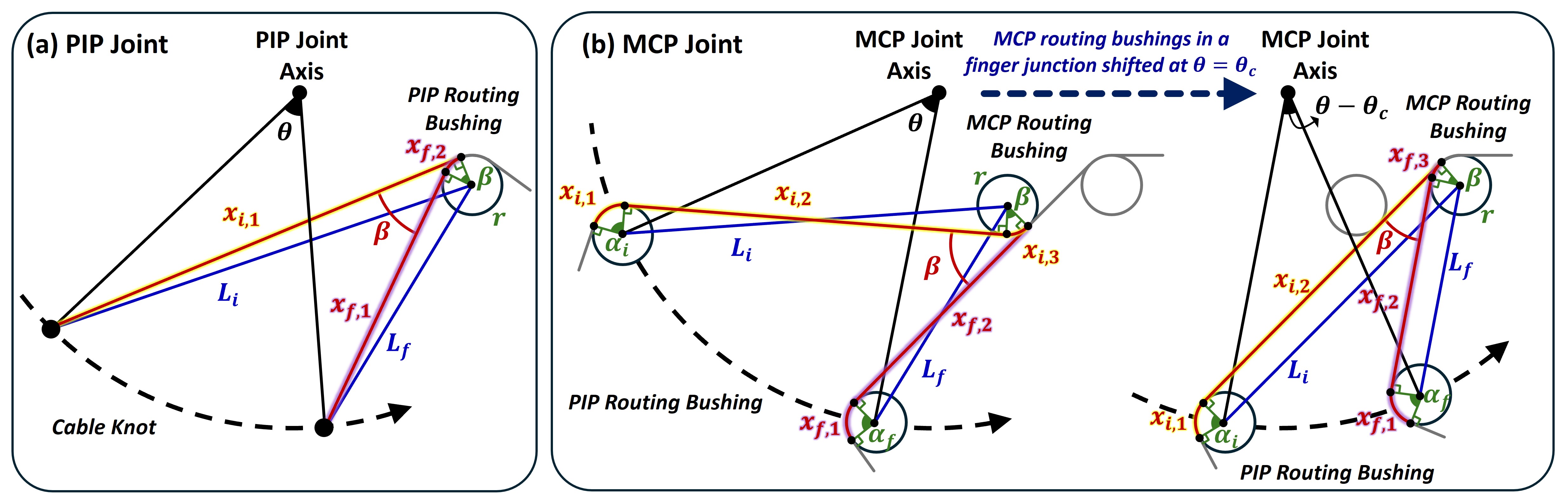}
\caption{Relationship between tendon length displacement and joint angles.
(a) PIP joint. (b) MCP joint.}
\label{fig_5}
\end{figure*}

Here, the tendon is guided by rotating bushings, resulting in a nonlinear tendon length displacement function $\ell(\theta_1,\theta_2)$, which can be computed from geometric configurations of the MCP and PIP joints, as illustrated in Fig.~\ref{fig_5}.
Let $x_i$ denote the initial tendon length between two phalanges and $x_f$ the tendon length after the joint rotates by $\theta$. 
The corresponding tendon length displacement can then be obtained from the difference between $x_i$ and $x_f$.
For the PIP joint, the tendon length displacement can be expressed as

\begin{align}
\ell_{\mathrm{pip}}(\theta) \
&= x_{i,1} - (x_{f,1} + x_{f,2}) \notag\\
&= \sqrt{L_i^2 - r^2} - \left( \sqrt{L_f^2 - r^2} + r \beta \right).
\label{eq:lpip}
\end{align}


\begin{table}[t]
\centering
\caption{Symbol Definition for Finger Modeling}
\label{tab:nomenclature}
\renewcommand{\arraystretch}{1.12}

{\footnotesize
*Subscripts 1 and 2 denote the MCP and PIP joints, respectively.
\par
}

\begin{tabular}{c|p{0.7\columnwidth}}
\hline
\textbf{Symbol} & \textbf{Description} \\ \hline

$\theta_1$, $\theta_2$ & MCP \& PIP joint angles. \\
$\theta_{\mathrm{i1}}$, $\theta_{\mathrm{i2}}$ & Initial offsets of the MCP \& PIP joint angles. \\
$\theta_{\mathrm{c}}$ & Switching angle between two bushings in MCP joint. \\
$\theta_{\mathrm{pre1}}$, $\theta_{\mathrm{pre2}}$ & Preloaded angles of the MCP \& PIP torsional springs. \\

$k_1$, $k_2$ & MCP \& PIP torsional spring constants. \\
$k_{\mathrm{sea}}$ & SEA spring constant. \\

$m_1$, $m_2$ & Masses of the proximal and distal phalanges. \\
$L_1$ & Position vector from MCP to PIP joint. \\
$L_{\mathrm{c1}}$, $L_{\mathrm{c2}}$ & Center-of-mass position vectors of the phalanges. \\

$R_{\mathrm{hand}}$ & Rotation matrix of the hand wrist. \\
$R_{\mathrm{j}}(\cdot)$ & Rotation matrix about a finger-joint axis. \\
$\hat{\mathbf{z}}$ & Unit vector along the gravitational $z$-direction. \\
$g$ & Gravitational acceleration. \\

$T$ & Tendon tension. \\
$x$ & Tendon excursion. \\
$\ell(\theta_1,\theta_2)$ & Tendon length displacement required to achieve the corresponding joint angles. \\
$x_i$, $x_f$ & Tendon length before and after joint rotation. \\

$L_i$, $L_f$ & Distances between the centers of the routing elements (cable knot or bushing) before and after joint rotation. \\

$\alpha_i$, $\alpha_f$ & Angle of the tendon arc in contact with the PIP routing bushing before and after joint rotation.\\
$\beta$ & Angle between tendon path segments between phalanges before and after joint rotation. \\
$r$ & Radius of the routing bushing. \\



\hline
\end{tabular}
\end{table}

For the MCP joint, two bushings are located at the finger junction, and the routing path changes at a specific switching angle $\theta_{\mathrm{c}}$. 
Therefore, the tendon length displacement must be computed separately for the two cases: $\theta < \theta_{\mathrm{c}}$ and $\theta \geq \theta_{\mathrm{c}}$. 
When $\theta < \theta_{\mathrm{c}}$, the tendon length displacement is given by

\begin{align}
\ell_{\mathrm{mcp1}}(\theta)
&= (x_{i,1} + x_{i,2} + x_{i,3}) - (x_{f,1} + x_{f,2}) \notag\\
&= \left( r \alpha_i + \sqrt{L_i^2 - 4r^2} + r \beta \right)
 - \left( r \alpha_f + \sqrt{L_f^2 - 4r^2} \right) \notag\\
&= \sqrt{L_i^2 - 4r^2} - \sqrt{L_f^2 - 4r^2} + r (2\beta - \theta),
\label{eq:lmcp1}
\end{align}

and when $\theta \geq \theta_{\mathrm{c}}$,

\begin{align}
\ell_{\mathrm{mcp2}}(\theta - \theta_{\mathrm{c}})
&= (x_{i,1} + x_{i,2}) - (x_{f,1} + x_{f,2} + x_{f,3}) \notag\\
&= \left( r \alpha_i + L_i \right)
 - \left( r \alpha_f + L_f + r \beta \right) \notag\\
&= L_i - L_f - r (\theta - \theta_{\mathrm{c}}).
\label{eq:lmcp2}
\end{align}

Summarizing the two cases, $\ell_{\mathrm{mcp}}(\theta)$ is expressed as

\begin{align}
\ell_{\mathrm{mcp}}(\theta) &=
\begin{cases}
\ell_{\mathrm{mcp1}}(\theta), & \text{if } \theta < \theta_{\mathrm{c}}, \\[4pt]
\ell_{\mathrm{mcp1}}(\theta_{\mathrm{c}}) + \ell_{\mathrm{mcp2}}(\theta - \theta_{\mathrm{c}}), & \text{if } \theta \ge \theta_{\mathrm{c}}.
\end{cases}
\label{eq:lmcp}
\end{align}

and the overall tendon length displacement $\ell(\theta_1,\theta_2)$ is obtained as the sum of the MCP and PIP contributions.

\begin{align}
\ell(\theta_1, \theta_2) = \ell_{\mathrm{mcp}}(\theta_1) + \ell_{\mathrm{pip}}(\theta_2)
\label{eq:ltotal}
\end{align}

The optimization problem, formulated as shown in \eqref{eq:opt}, aims to minimize the total energy of the system at certain tendon excursion. 
The tendon excursion $x$ was discretized into small steps, and at each step, the state with the minimum energy was selected as the next step’s state. Consequently, the only decision variables were the two joint angles, since the SEA spring displacement---and thus the tension---could be readily obtained by $\Delta x_{\mathrm{sb}} = (x - \ell(\theta_1, \theta_2))/2$. The constraints were defined by the joint range of motion, expressed as $\mathrm{RoM}_j$, and by the elastic limits of the SEA spring, expressed as $x_{\mathrm{sb,max}}$. This nonlinear optimization problem was solved in MATLAB using the sequential quadratic programming (SQP) algorithm. Consequently, the MCP and PIP joint angles, $\theta_1(x)$ and $\theta_2(x)$, as well as the corresponding tendon tension $T(x)$, can be expressed as functions of the tendon excursion $x$. The profiles of modeling results are presented together with the experimental results in Section~\ref{sec:finger_actuation}.

\begin{align}
\min_{\theta_1,\,\theta_2} \quad & E_{\mathrm{Total}}(\theta_1,\theta_2) \notag\\
\text{s.t.}\quad
& 0 \leq \theta_{\mathrm{i}j} + \theta_j \leq \mathrm{RoM}_j,\quad j\in\{1,2\}, \notag\\
& 0 \leq \frac{x - \ell(\theta_1,\theta_2)}{2} \leq x_{\mathrm{sb,max}}
\label{eq:opt}
\end{align}

\subsection{Tension Profile Characterization}

\begin{figure}[t]
\centering
\includegraphics[width=0.95\columnwidth]{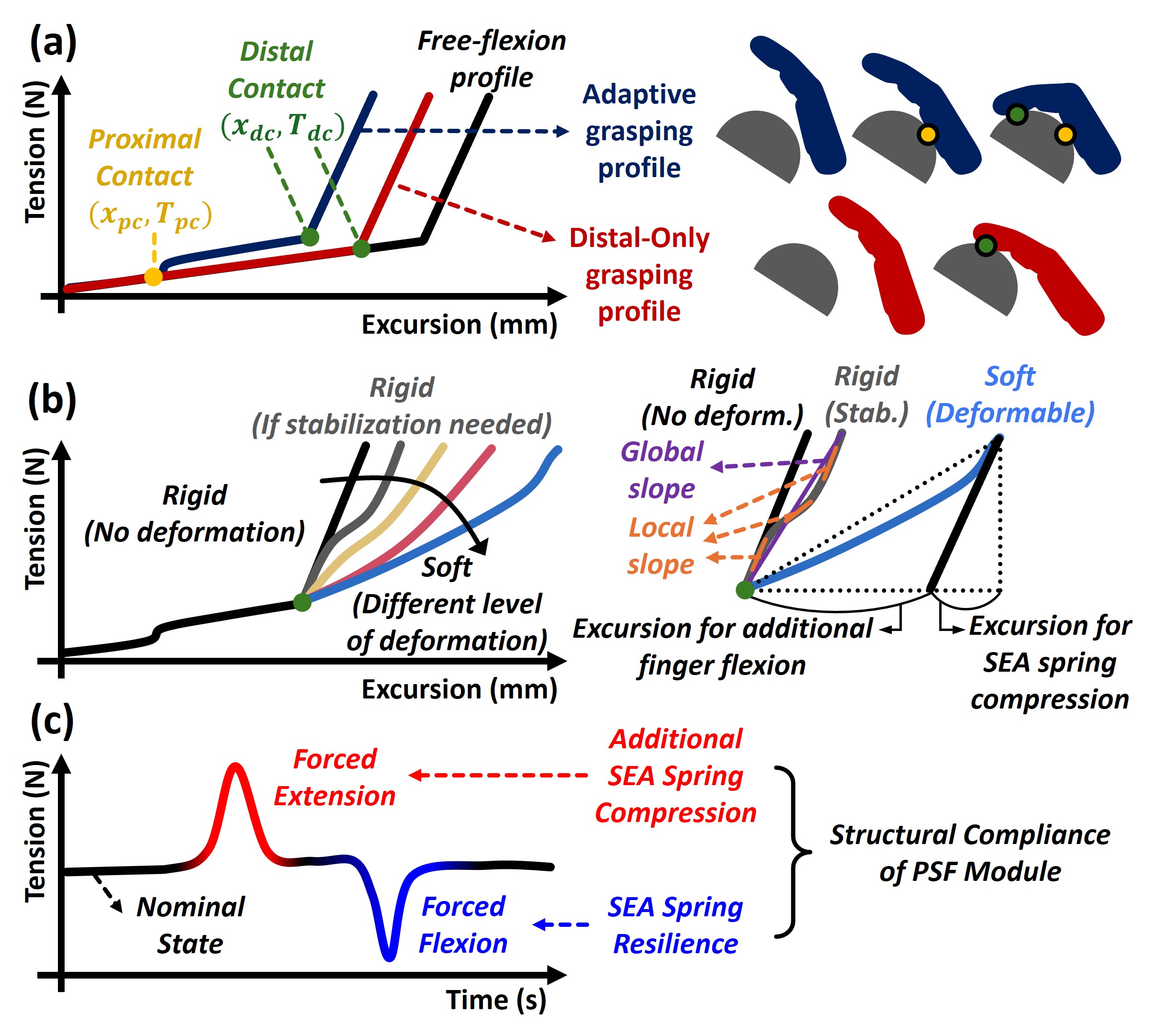}
\caption{Tension profile characterization.
(a) Tension profiles reflecting contact events for different grasp types. 
(b) Tension profile reflecting object deformation and relative stiffness. 
(c) Tension profile reflecting finger configuration changes by external disturbances.}
\label{fig_6}
\end{figure}

Fig.~\ref{fig_6}(a) illustrates the profiles of finger--object contact events for different grasp types. In an underactuated finger, the proximal and distal phalanges can sequentially contact the object, allowing the finger to adaptively wrap around it, or contact can occur only at the distal phalanx~\cite{spiers_single-grasp_2016}. Contact is indicated by deviation from the free-flexion reference. Proximal contact constrains MCP joint motion, preventing the finger from following its original minimum-energy trajectory, which results in a slight upward shift of the tension profile. Distal contact halts finger motion, leaving only the SEA spring to compress and producing a sharp slope increase with an inflection point. The corresponding contact excursion can then be mapped to the joint angles.
Let $x_{\mathrm{pc}}$ and $x_{\mathrm{dc}}$ denote the tendon excursions at proximal and distal contact, respectively, and $T_{\mathrm{pc}}$ and $T_{\mathrm{dc}}$ the corresponding tendon tensions.
After proximal contact, the subsequent tendon excursion contributes only to the PIP joint motion and SEA spring compression; therefore, tendon excursion contributing to the PIP joint motion can be calculated as

\begin{equation}
\Delta x_{\mathrm{pip}} = x_{\mathrm{dc}} - x_{\mathrm{pc}} 
- \frac{T_{\mathrm{dc}} - T_{\mathrm{pc}}}{k_{\mathrm{sea}}}.
\label{eq:xpip}
\end{equation}




The PIP joint angle change after proximal contact can be directly obtained from the inverse of the tendon length function $\ell_{\mathrm{pip}}^{-1}(\cdot)$ derived in~\eqref{eq:lpip}, which was approximated by a five-order polynomial to enable fast computation of the inverse function.
Since $\theta_1$ and $\theta_2$ are modeled as functions of tendon excursion, the estimated joint angles after distal contact are given by

\begin{equation}
\begin{cases}
\theta_{1,\mathrm{est}} = \theta_1(x_{\mathrm{pc}}),\\[3pt]
\theta_{2,\mathrm{est}} = \theta_2(x_{\mathrm{pc}}) + \ell_{\mathrm{pip}}^{-1}\!\left(\Delta x_{\mathrm{pip}}\right).
\end{cases}
\label{eq:joint_angle_contact}
\end{equation}


Fig.~\ref{fig_6}(b) presents tension profiles during the loading phase after distal contact, depending on the level of object deformation. For rigid objects, the slope theoretically matches the SEA spring constant, whereas for soft objects, the finger continues to flex even after distal contact, yielding a reduced slope with object deformation. Thus, the slope indicates the portion of the tendon excursion contributing to object deformation and provides insight into object stiffness. Both local slope, which reflects instantaneous changes, and global slope, measured from the contact excursion, can be considered. Even for rigid objects, slight motion may occur until the object’s configuration stabilizes due to inter-finger force equilibrium. Consequently, the global slope is suitable when evaluating the overall degree of deformation, while the local slope is more desirable for identifying intervals exhibiting rigid-like stiffness. Although the absolute value of object stiffness is not directly obtainable due to the indeterminate contact location, rigid and soft objects can be clearly distinguished, and relative stiffness comparisons remain valid within the same configuration.

Fig.~\ref{fig_6}(c) shows tension profiles under externally induced finger configuration changes when a finger is partially flexed or holding an object. 
Forcibly extending the finger compresses the SEA spring further and increases tension, while forcibly flexing the finger or losing support from the object being held releases the spring and reduces tension.
These responses demonstrate that the PSF module can detect external disturbances through its structural compliance: flexion compliance is inherent to tendon-driven mechanisms, whereas extension compliance is uniquely enabled by the SEA.

\subsection{Grasping Strategy and Algorithm for State Estimation}

\begin{figure}[t]
\centering
\includegraphics[width=\columnwidth]{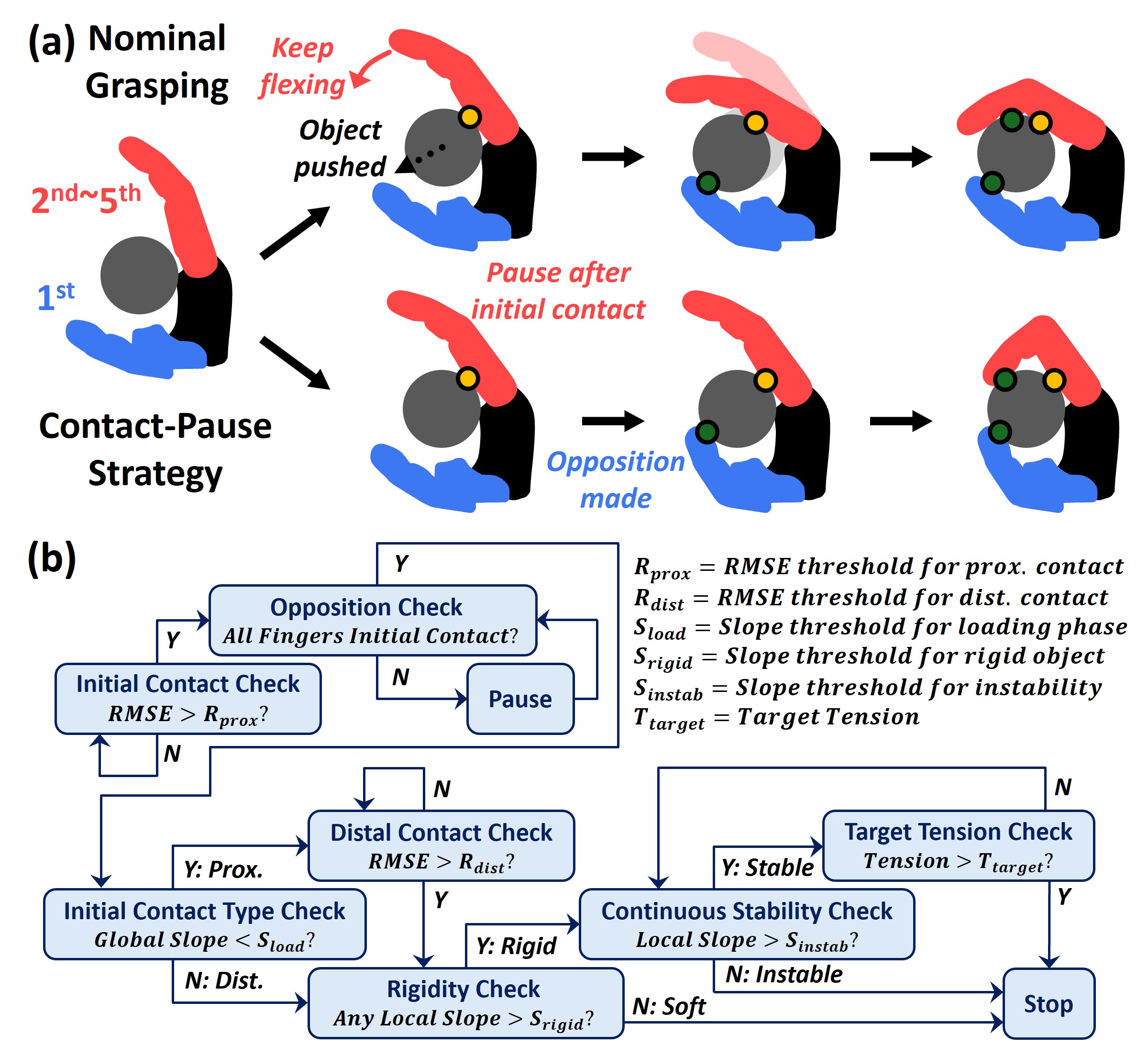}
\caption{Grasping strategy and algorithm.
(a) Contact-pause strategy to prevent object displacement and maintain the initial location.
(b) Rule-based grasping algorithm.}
\label{fig_7}
\end{figure}

When estimating joint angles from contact detection and relative stiffness from slope, it is critical to constrain the object, which is typically achieved through thumb opposition. If a finger contacts the object before opposition, object pushing may occur as in the nominal grasping case in Fig.~\ref{fig_7}(a), displacing the object and introducing errors in estimation. To prevent this, a strategy similar to that of~\cite{jentoft_limits_2014} was adopted: the first contacting finger is paused, and actuation resumes only after all fingers have made contact. This contact-pause strategy ensures that the object retains its initial position.

Contact detection from tension profile deviations is implemented using a sliding-window root-mean-square error (RMSE) method. The control loop operates at \SI{1}{\kilo\hertz}, and RMSE is computed over a 50-sample window (\SI{50}{\milli\second}). Before computation, the window is offset-corrected by aligning its first sample to the corresponding reference value, ensuring consistent comparison with the reference profile. When the RMSE exceeds a predefined threshold $R_{\mathrm{prox}}$, contact is detected and assigned to the midpoint of the window. To identify the initial contact type, the global slope is evaluated \SI{75}{\milli\second} after threshold crossing: values below $S_{\mathrm{load}}$ indicate proximal contact, while values above $S_{\mathrm{load}}$ indicate distal contact. If the initial contact is distal, the system proceeds directly to rigidity evaluation; if proximal, an additional RMSE-based search with threshold $R_{\mathrm{dist}}$ is conducted to detect distal contact before rigidity evaluation. The object is classified as rigid if any local slope, computed over a sliding window, exceeds $S_{\mathrm{rigid}}$; otherwise, it is deemed soft, and actuation is halted to regulate the grasping force, thereby preventing excessive deformation or fracture. To prevent misclassification or instability due to object stabilization, actuation is also terminated if the local slope falls below $S_{\mathrm{instab}}$ during loading. This rule-based strategy enables accurate estimation and stable grasp execution. Fig.~\ref{fig_7}(b) illustrates the entire grasping algorithm implemented for each finger.

\section{Experimental Validations}

\begin{figure}[t]
\centering
\includegraphics[width=0.9\columnwidth]{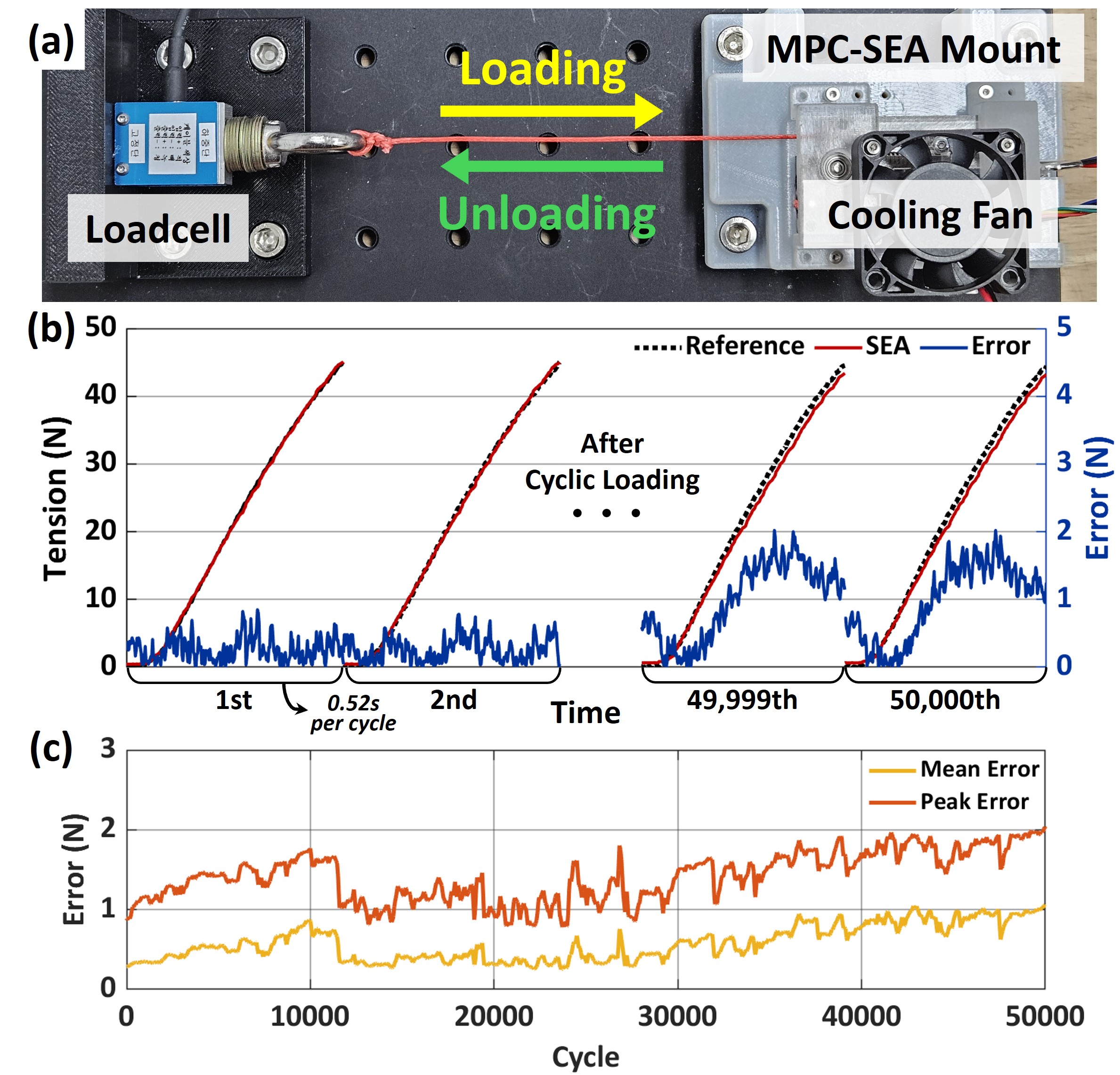}
\caption{Results of cyclic loading test of MPC-SEA. (a) Experiment setup. (b) Tension profiles over time for the first and last two cycles. (c) Mean error and peak error over cycles.}
\label{fig_8}
\end{figure}

\subsection{MPC-SEA Experiment}
To evaluate the accuracy and reliability of the MPC-SEA’s tension sensing, a cyclic loading test was conducted. As shown in Fig.~\ref{fig_8}(a), the cable of a fixed MPC-SEA was connected to a load cell (333FDX, KTOYO, Korea). Each cycle consisted of loading from \SI{0}{\newton} to \SI{45}{\newton} ($\approx$90\% of maximum tension) and then unloading back to \SI{0}{\newton}. The tension was recorded over loading, which is relevant to grasping, and compared with the reference tension measured by the load cell.
Before the test, a MPC-SEA was calibrated to compensate for slight parameter variations among components, such as springs and potentiometers, as well as minor offsets introduced by the potentiometer’s placement during assembly.
The calibration was performed using data from the 100 cycles, which were fit with a first-order linear model via the least squares method to determine the slope and intercept of \eqref{eq:sea3}.
After the calibration process, a total of 50,000 cycles was performed.

Fig.~\ref{fig_8}(b) shows the tension profiles over time for the first and last two cycles. In the first two cycles, the error remained below \SI{0.9}{\newton} across the entire range, indicating high accuracy. In the last two cycles, the error increased slightly in the higher-tension range, with a peak error of about \SI{2}{\newton}. However, considering 50,000 cycles of severe repetitive loading, the sensing performance was well preserved. Errors mainly arose from transmission friction, spring nonlinearities, and misalignment between the sliding block and the potentiometer due to assembly tolerances.
Fig.~\ref{fig_8}(c) shows the mean error and peak error, computed over all timesteps in each cycle. Although both error metrics increased compared to the initial cycles as the number of cycles approached 50,000, the mean error remained below \SI{1}{\newton} and the peak error below \SI{2}{\newton} throughout the entire test. These results confirm that the MPC-SEA maintained accurate sensing performance even after 50,000 cycles, demonstrating the reliability of its structure and mechanism. 

\subsection{Finger Actuation Experiment}

\begin{figure*}[t]
\centering
\includegraphics[width=0.88\textwidth]{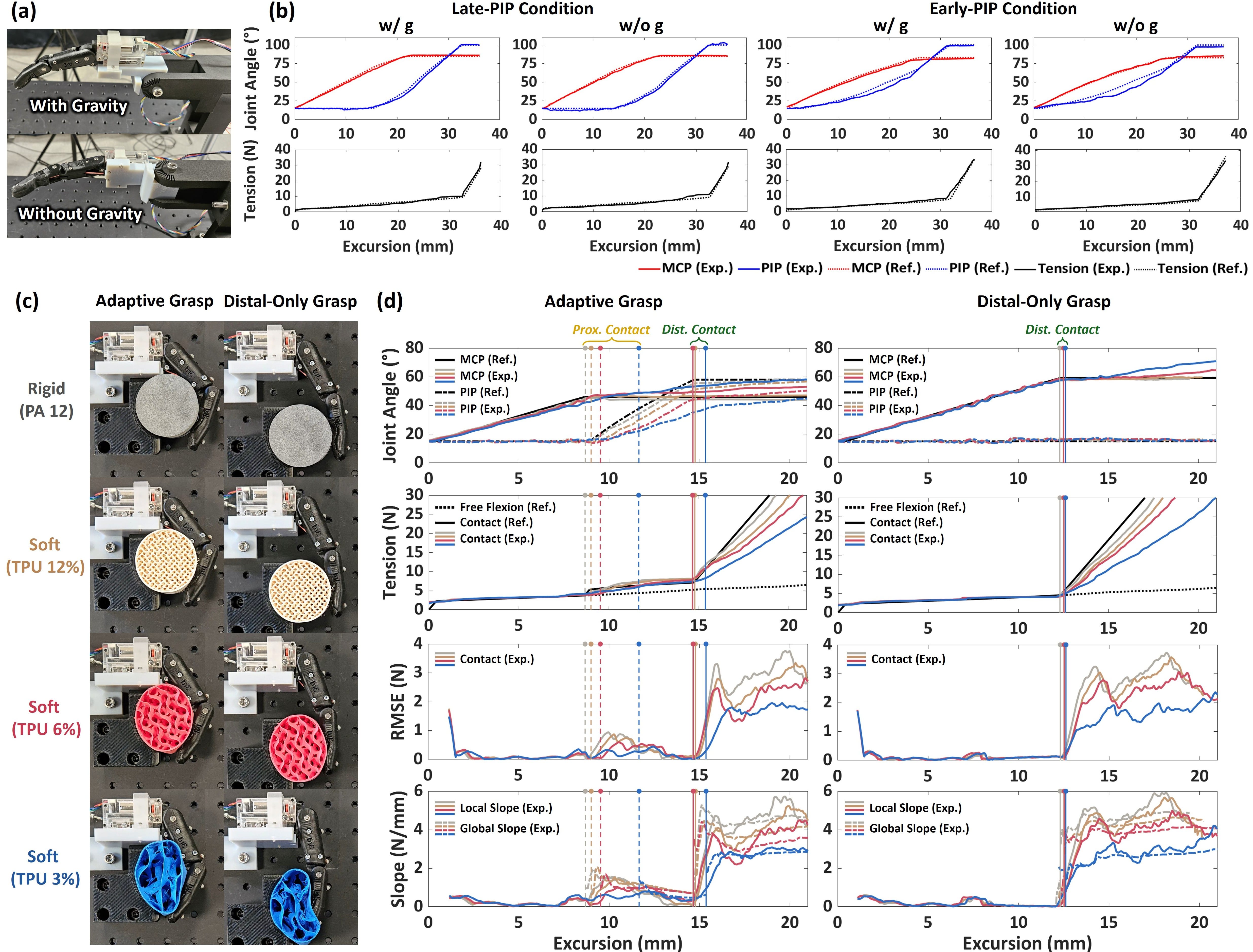}
\caption{Results of finger actuation experiments
(a) Experimental setup for free-flexion cases.
(b) Free-flexion results for joint angles and tension.
(c) Experimental setup for finger--object contact cases.
(d) Finger-object contact results for joint angles, tension, RMSE, and slope.}
\label{fig_9}
\end{figure*}

\label{sec:finger_actuation}
\subsubsection{Free-Flexion Case}
To validate the accuracy of the modeling, experiments were conducted using a PSF module. Fig.~\ref{fig_9}(a) shows the setup for the free-flexion cases. Two representative orientations were selected: one with the palm facing sideways, unaffected by gravity, and another with the palm facing downward, subject to gravity. The torsional spring preloads of the MCP and PIP joints were also adjusted in two combinations to examine different actuation sequences: a late-PIP condition (MCP \SI{36}{\degree}/PIP \SI{32}{\degree}) and an early-PIP condition (MCP \SI{28}{\degree}/PIP \SI{12}{\degree}). The actuation was performed by applying the maximum PWM command, resulting in a closing time of \SI{1.3}{\second} at the highest motor speed. The real-time tension was sensed through the SEA, while the actual joint angles were measured with a motion capture system (OptiTrack, NaturalPoint Inc., USA). Two reflective markers were attached to each phalanx, and joint angles were computed from changes in marker vectors between initial and final configurations.

Each experiment was repeated five times, and the average results are shown in Fig.~\ref{fig_9}(b). Under the late-PIP condition, the joint angles and tension matched the modeling reference with high accuracy. Under the early-PIP condition, slight discrepancies were observed in the PIP joint, likely because the weaker preload caused the two joints to actuate more simultaneously, making the motion more susceptible to friction. In both conditions, gravitational effects were minimal, as the finger was lightweight with relatively high joint stiffness, making spring forces dominant. To prevent ejection in underactuated fingers and to secure a larger workspace, the late-PIP condition was selected as the final parameter.
Although the model is based on a quasi-static assumption, the experimental results demonstrate that it remains valid even under dynamic actuation. 
Tendon-based proprioception allows for a simple finger structure without additional sensors, yielding a lightweight finger of only about \SI{28.6}{\gram}; thus, inertial effects are negligible, and the quasi-static assumption remains well justified.

\subsubsection{Object Contact Case}
To validate the plausibility of the tension-profile characterization, the experimental setup for finger–object contact cases was configured as shown in Fig.~\ref{fig_9}(c).
Both adaptive and distal-only grasp types were considered, with a fixture substituting for the role of thumb opposition. 
The test object was a \SI{50}{\milli\meter} diameter cylinder; the rigid one was 3D-printed in PA12 (3D HR PA12, HP Inc., USA), and the soft ones were printed in TPU (VarioShore TPU, colorFabb, Netherlands) with varying infill densities. 
A gyroid infill pattern, which is nearly isotropic~\cite{bean_numerical_2022}, was selected. For this experiment, actuation was not halted as in Fig.~\ref{fig_7}(b) but instead uniformly increased to \SI{40}{\newton} to observe slope variations by deformation during loading.
The RMSE thresholds for contact detection were set to $R_{\mathrm{prox}} = \SI{0.4}{\newton}$ and $R_{\mathrm{dist}} = \SI{1.0}{\newton}$, with the loading slope threshold set to $S_{\mathrm{load}} = \SI{2}{\newton\per\milli\metre}$.

Fig.~\ref{fig_9}(d) presents the modeled reference and experimental joint angles and tension profiles with RMSE and slopes, and Table~\ref{tab:angle_stiffness} summarizes the estimated joint angles with the actual joint angles measurements and the relative stiffness values for different grasp conditions.
The reference profiles for joint angles and tension were generated for the case of a rigid object, obtained by solving the optimization problem in \eqref{eq:opt} while constraining the $\mathrm{RoM}_j$ to the angles expected at object contact.
The profiles clearly exhibited the distinct proximal and distal contact patterns characterized in Fig.~\ref{fig_6}(a), and the slopes varied noticeably with stiffness level as in Fig.~\ref{fig_6}(b).
In both grasp types, the slope for the rigid object was slightly lower than the reference, likely due to deformation of the silicone fingertip. 
Since this deformation consistently occurs at every contact, the experimentally measured rigid slope can serve as a practical baseline.
Between the two slope metrics, the local and global slopes exhibited similar tendencies, but the global slope proved more stable and reliable for distinguishing relative stiffness levels.

\begin{table*}[t]
\centering
\caption{Joint Angle and Relative Stiffness Estimation under Different Grasp Conditions}
\label{tab:angle_stiffness}
\renewcommand{\arraystretch}{1.1}
\begin{tabular}{>{\centering\arraybackslash}m{3.0cm}|
                >{\centering\arraybackslash}m{1.5cm}|
                cccc|cccc}
\hline
\textbf{Parameter} & \textbf{Symbol} & \multicolumn{4}{c|}{\textbf{Adaptive Grasp}} & \multicolumn{4}{c}{\textbf{Distal-Only Grasp}} \\ \cline{3-10}
 &  & Rigid & Soft 12\% & Soft 6\% & Soft 3\% & Rigid & Soft 12\% & Soft 6\% & Soft 3\% \\ \hline

\multirow{3}{*}{\centering \textbf{MCP Joint Angle}} 
 & $\theta_{1,\mathrm{est}}$ [\SI{}{\degree}] & 46.10 & 47.28 & 49.13 & 56.77 & 58.20 & 58.76 & 58.84 & 58.68 \\
 & $\theta_{1,\mathrm{act}}$ [\SI{}{\degree}] & 44.20 & 46.49 & 49.30 & 53.82 & 57.64 & 58.31 & 58.48 & 57.10 \\
 & error$_1$ [\SI{}{\degree}] & 1.90 & 0.79 & -0.17 & 2.95 & 0.56 & 0.45 & 0.36 & 1.58 \\ \hline

\multirow{3}{*}{\centering \textbf{PIP Joint Angle}} 
 & $\theta_{2,\mathrm{est}}$ [\SI{}{\degree}] & 56.42 & 55.12 & 51.47 & 43.44 & 15.07 & 15.07 & 15.07 & 15.07 \\
 & $\theta_{2,\mathrm{act}}$ [\SI{}{\degree}] & 54.80 & 50.92 & 43.91 & 37.54 & 16.17 & 15.88 & 16.10 & 15.90 \\
 & error$_2$ [\SI{}{\degree}] & 1.62 & 4.20 & 7.56 & 5.90 & -1.10 & -0.81 & -1.03 & -0.83 \\ \hline

\textbf{Relative Object Stiffness} 
 & $k_{\mathrm{rel}}$ [N/mm] & 4.26 & 3.63 & 2.96 & 2.65 & 4.27 & 3.95 & 3.75 & 2.45 \\ \hline

\end{tabular}
\end{table*}

In the adaptive grasp, rigid objects caused the MCP angle to remain fixed after proximal contact, whereas soft objects allowed the MCP angle to continue increasing slightly along with the PIP joint.
Namely, tendon excursion that would have contributed solely to PIP flexion after proximal contact in the rigid object case also contributes to MCP flexion when interacting with soft objects, making joint angle estimation more challenging for soft objects.
Nevertheless, the rule-based contact detection scheme naturally accounts for such interaction.
Since the weak resistance of soft objects produces a smaller deviation from the reference profile, it delays the proximal threshold ($R_{\mathrm{prox}}$) crossing. 
These results yielded larger estimated MCP angles and smaller PIP angles, thereby mitigating potential errors and preserving consistency with the actual grasp dynamics.
As shown in Table~\ref{tab:angle_stiffness}, the MCP joint-angle estimation errors were all within \SI{3}{\degree}. In contrast, the PIP estimation errors were relatively larger for softer objects, mainly because the deviation at distal contact is smaller than in the rigid object case, which delays the distal threshold ($R_{\mathrm{dist}}$) crossing and leads to slightly overestimated PIP angles.
While the estimation scheme is primarily tailored to rigid object behavior, it nonetheless maintains acceptable performance when applied to soft objects as well.

Meanwhile, in the distal-only grasp, where the interaction is simpler than in the adaptive grasp, the contact excursion remained nearly unchanged across different object stiffnesses. 
Therefore, the estimated joint angles remained similar regardless of object stiffness, and the estimation errors were consistently very small, indicating that accurate joint angle estimation was achieved.

\begin{figure}[t]
\centering
\includegraphics[width=0.88\columnwidth]{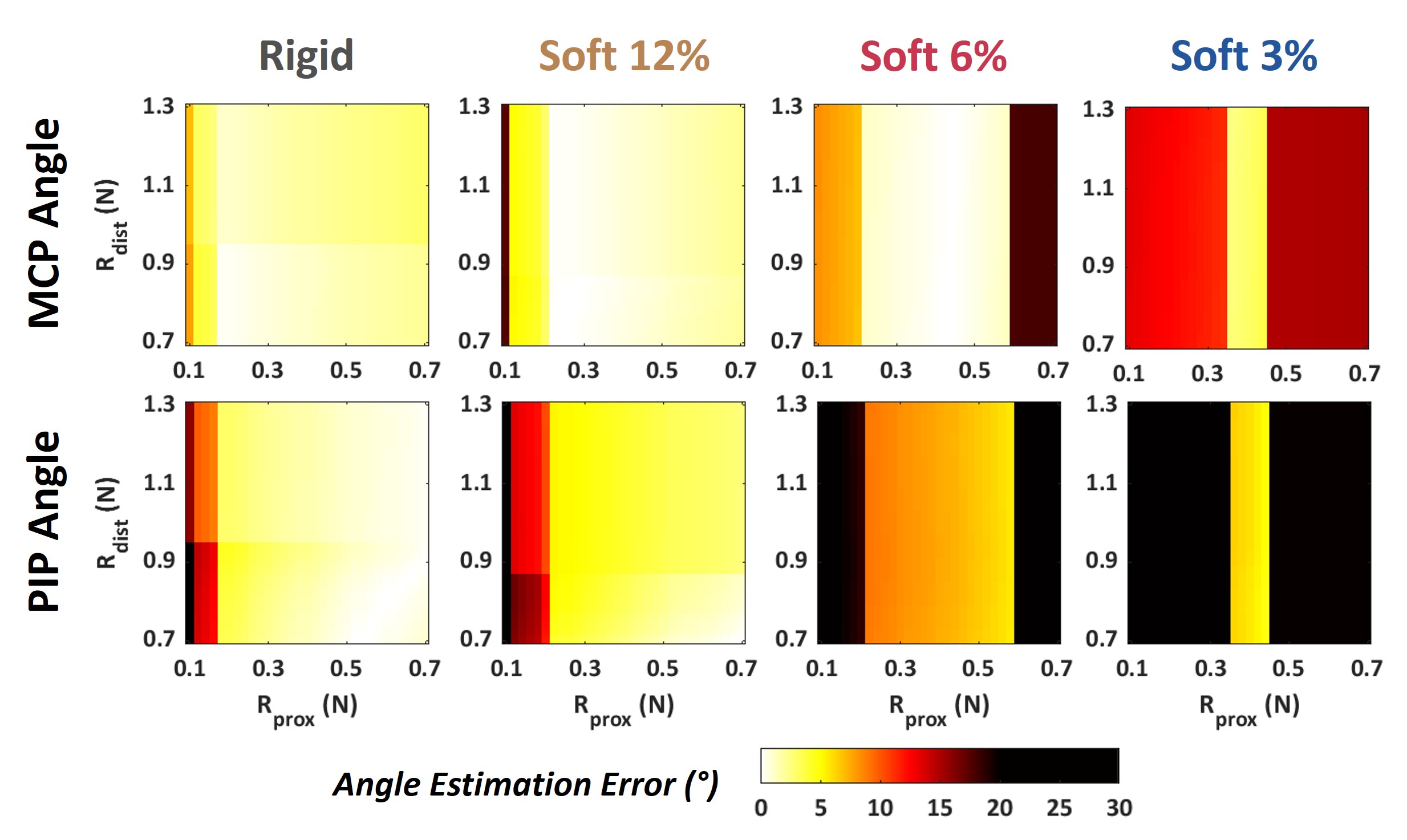}
\caption{Heatmap analysis of joint angle estimation errors across threshold $R_{\mathrm{prox}}$ and $R_{\mathrm{dist}}$.}
\label{fig_10}
\end{figure}

In addition, to examine the influence of the threshold value selection in the rule-based algorithm, the joint angle estimation errors were evaluated while varying $R_{\mathrm{prox}}$ and $R_{\mathrm{dist}}$ within the ranges 0.1--0.7\SI{}{\newton} and 0.7--1.3\SI{}{\newton}, respectively, as shown in Fig.~\ref{fig_10}. During this analysis, $S_{\mathrm{load}}$ was fixed at \SI{2}{N/mm}, since this threshold was found to be insensitive; varying it between 1--3 did not affect the estimation results.
Fig.~\ref{fig_10} shows that the PIP angle is generally more sensitive to threshold selection than the MCP angle. 
When $R_{\mathrm{prox}}$ is too small, the estimator becomes sensitive to noise and may falsely detect contact during free flexion. In contrast, if $R_{\mathrm{prox}}$ is set too high, the proximal contact deviation for soft objects may not reach the threshold, increasing estimation error.
$R_{\mathrm{dist}}$ has only a minor influence, as the distal contact produces a clear inflection in the tension profile. 
Sensitivity also increases with softer objects, whose tension profiles deviate more from the modeled reference. This effect is most pronounced for the soft object of 3\% infill, where improper threshold selection can lead to noticeable errors.
These observations collectively indicate that while the proposed estimation scheme is broadly robust, careful selection of $R_{\mathrm{prox}}$ is important, particularly for highly soft objects.

\section{Grasp Functionality Demonstration}
\subsection{Grasp Posture Reconstruction}

\begin{figure}[t]
\centering
\includegraphics[width=0.85\columnwidth]{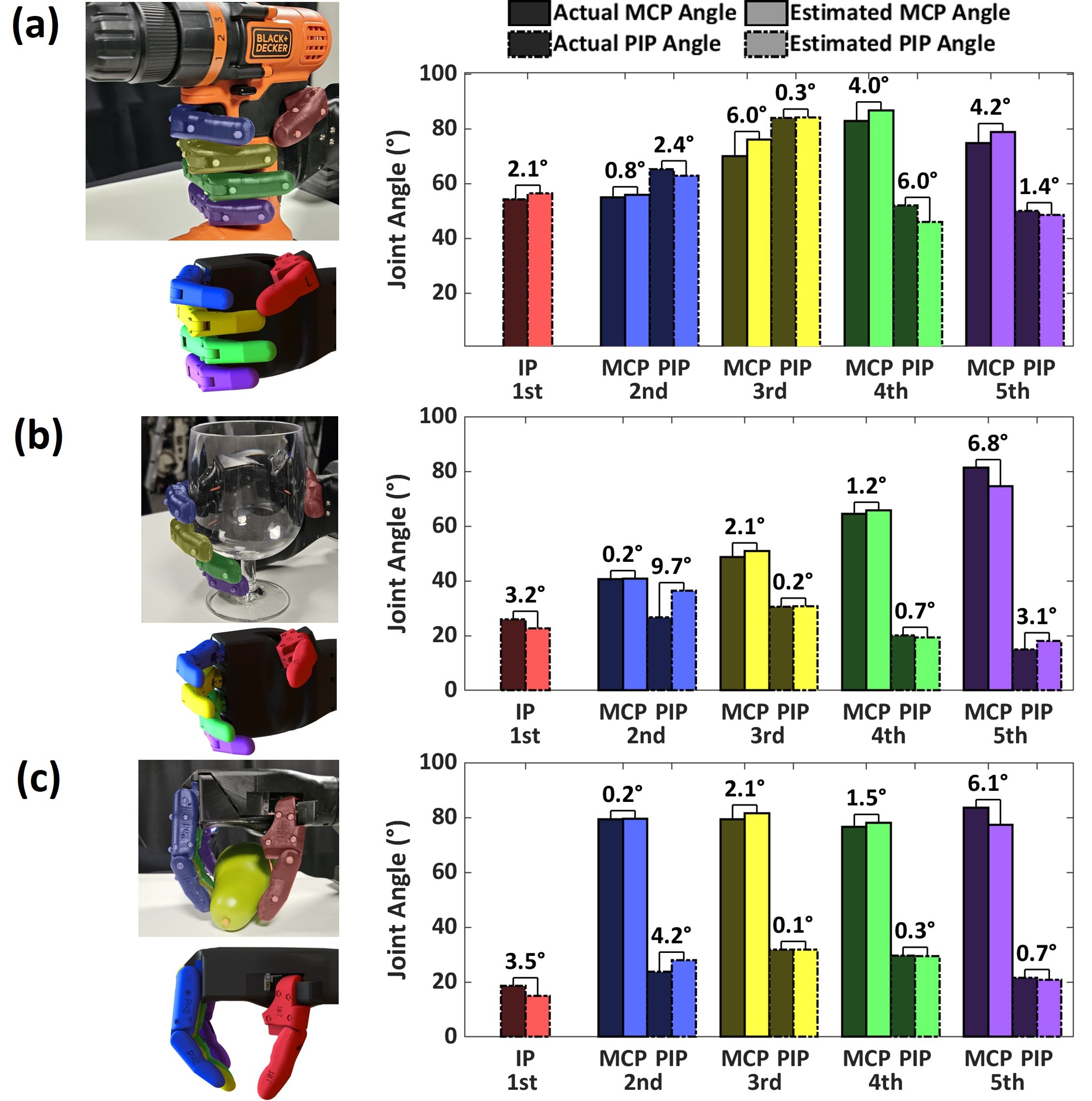}
\caption{Comparison of the actual and estimated grasp posture for different objects from YCB object set.
(a) Hand drill. (b) Wine glass. (c) Plastic pear.}
\label{fig_11}
\end{figure}

\begin{figure}[t]
\centering
\includegraphics[width=0.8\columnwidth]{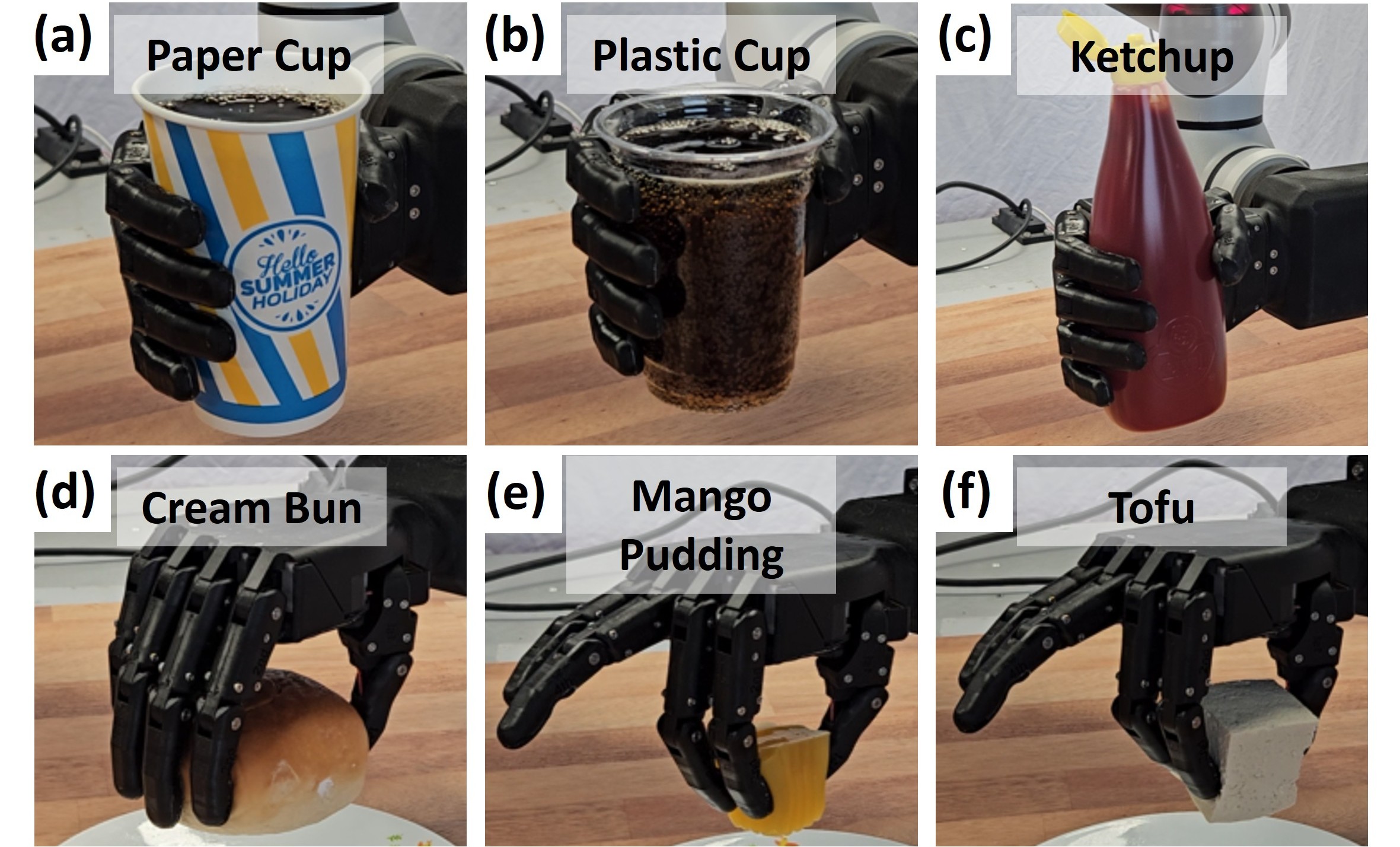}
\caption{Safe grasp force regulation for deformable objects.
(a) Paper cup. (b) Plastic cup. (c) Ketchup. (d) Cream bun. (e) Mango pudding. (f) Tofu.}
\label{fig_12}
\end{figure}

The joint angle estimation of individual fingers was extended to reconstruct the overall grasp posture. Three irregularly shaped YCB objects~\cite{calli_ycb_2015} were selected: a power drill, a wine glass, and a plastic pear. The drill and glass correspond to adaptive grasps, while the pear corresponds to a distal-only grasp. 
Actual joint angles were measured using the same motion capture procedure described in Section~\ref{sec:finger_actuation}.
As shown in Fig.~\ref{fig_11}, the estimated posture from tension profiles was visualized on a CAD model and compared with the actual posture. The average error per joint angle was \SI{3.0}{\degree} for the drill and glass and \SI{2.1}{\degree} for the pear, indicating high overall accuracy.

\subsection{Safe Grasp Force Regulation for Deformable Objects}
With rigidity evaluation in the grasping algorithm, deformable object grasping was demonstrated on common objects, including a paper cup, plastic cup, ketchup, cream bun, mango pudding, and tofu, as shown in Fig.~\ref{fig_12}. The first four objects were grasped with all five fingers, while the last two smaller ones were grasped using a tripod pinch. In all cases, finger flexion was halted appropriately after distal contact, preventing excessive deformation or fracture while enabling stable lifting.

\begin{figure}[t]
\centering
\includegraphics[width=0.9\columnwidth]{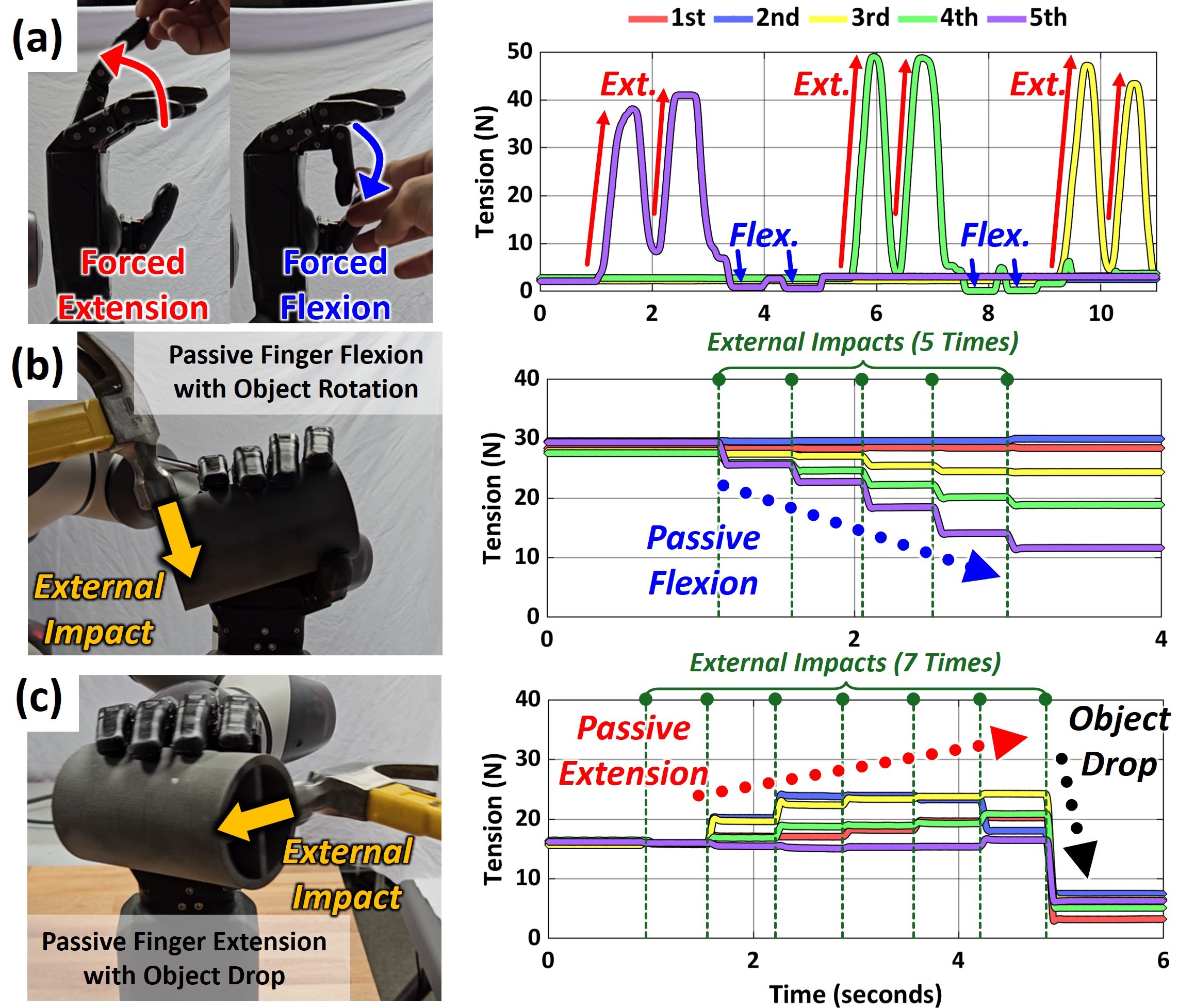}
\caption{Disturbance detection with SEA-driven compliance.
(a) Tension change during forced flexion and extension of an initially flexed finger.
(b), (c) Passive finger flexion and extension caused by external impacts.}
\label{fig_13}
\end{figure}

\subsection{Disturbance Detection with Compliance}
Building on the characterization in Fig.~\ref{fig_6}(c), disturbance detection through structural compliance was experimentally verified. As shown in Fig.~\ref{fig_13}(a), when a finger preloaded to about \SI{2.5}{\newton} was forcibly extended, the SEA spring compressed further with increased tension, whereas forced flexion caused the spring to release with decreased tension. Fig.~\ref{fig_13}(b) and (c) show responses to hammer strikes on a grasped cylinder: in (b), ulnar-side rotation of the object induced additional flexion of the 3rd--5th fingers, whereas in (c), repeated impacts gradually extended the fingers until the object was released, accompanied by a sharp tension drop.
These results demonstrate that the system can detect the nature of external disturbances through tension changes, while compliance simultaneously enables the fingers to maintain contact without damage, even under external impacts such as hammer strikes that alter the object’s configuration.

\subsection{Blind Grasping with Object Recognition}

\begin{figure}[t]
\centering
\includegraphics[width=0.85\columnwidth]{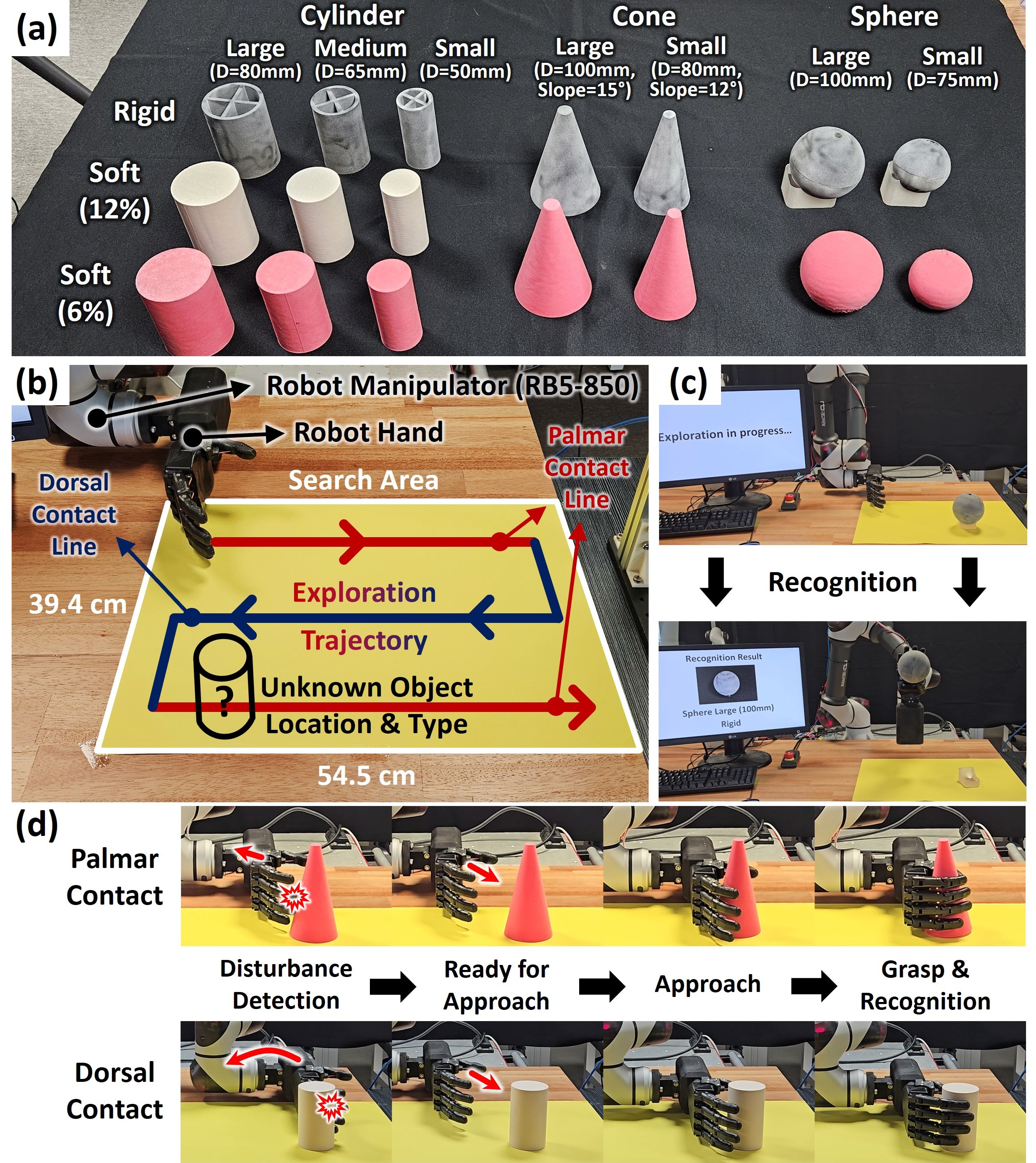}
\caption{Blind grasping with object recognition demonstration.
(a) Sample Objects with varying size, shape, and stiffness.
(b) Task setup with exploration trajectory.
(c) Task completion with object recognition result.
(d) Execution of approach and grasp after palmar and dorsal contact.}
\label{fig_14}
\end{figure}

A blind grasping task with object recognition was conducted to demonstrate the integrated proprioceptive functionality. 
This task was inspired by the way humans sweep their hands across a table to locate and grasp objects without vision. 
The demonstration integrates three key capabilities: object localization through disturbance detection, autonomous grasp force regulation for safely handling soft objects, and object recognition based on grasp posture reconstruction and stiffness classification.
As shown in Fig.~\ref{fig_14}(b), the hand mounted on a robot manipulator (RB5-850, Rainbow Robotics, Korea) explored a defined search area by executing three consecutive sweeps—palmar, dorsal, and palmar. 
Before exploration, the fingers were flexed slightly for \SI{50}{\milli\second} to preload tension, thereby enabling bidirectional disturbance detection. 
Once contact was detected, the hand executed a predefined approach and completed the grasp, with dorsal contacts leading to an approach from behind the object as shown in Fig.~\ref{fig_14}(d).

Seventeen objects with varying size, shape, and stiffness were used, including cylinders, cones, and spheres in rigid and soft forms, as shown in Fig.~\ref{fig_14}(a). All objects were fabricated with the same material as in Fig.~\ref{fig_9}(c). To prevent rolling or slipping during exploration, spheres were mounted on support bases, and soft objects had sandpaper attached to their bottoms.
For object recognition, a simple k-nearest neighbors (k-NN) model ($k=3$, Euclidean distance) was employed as a proof-of-concept demonstration. 
Object geometry (size and shape) was recognized using a kinematics-based approach that utilizes grasp posture~\cite{boruah_shape_2023, vasquez_proprioceptive_2017}, while stiffness classification was subsequently performed within each geometry category by comparing the average relative stiffness across five fingers. 
The feature vector of the training dataset consisted of nine joint angles and the averaged relative stiffness, collected from ten grasp trials per object. To enhance robustness against variability in the approach motion, the object positions were slightly varied during training.
As depicted in Fig.~\ref{fig_14}(c), the system successfully identified unknown objects and displayed the recognition result, with the complete process provided in the supplementary video.

The confusion matrices in Fig.~\ref{fig_15} show the recognition performance with 5-fold cross-validation. Geometry classification achieved an accuracy of 95.3\%. 
Stiffness classification for cones and spheres also showed high accuracy, whereas cylinders exhibited slightly lower performance, achieving around 80\% accuracy.
Nevertheless, the overall demonstration results confirm that the proposed estimation scheme remains effective in identifying both object geometry and stiffness using only tendon-based proprioception, without reliance on vision or tactile sensors.

\begin{figure}[t]
\centering
\includegraphics[width=0.9\columnwidth]{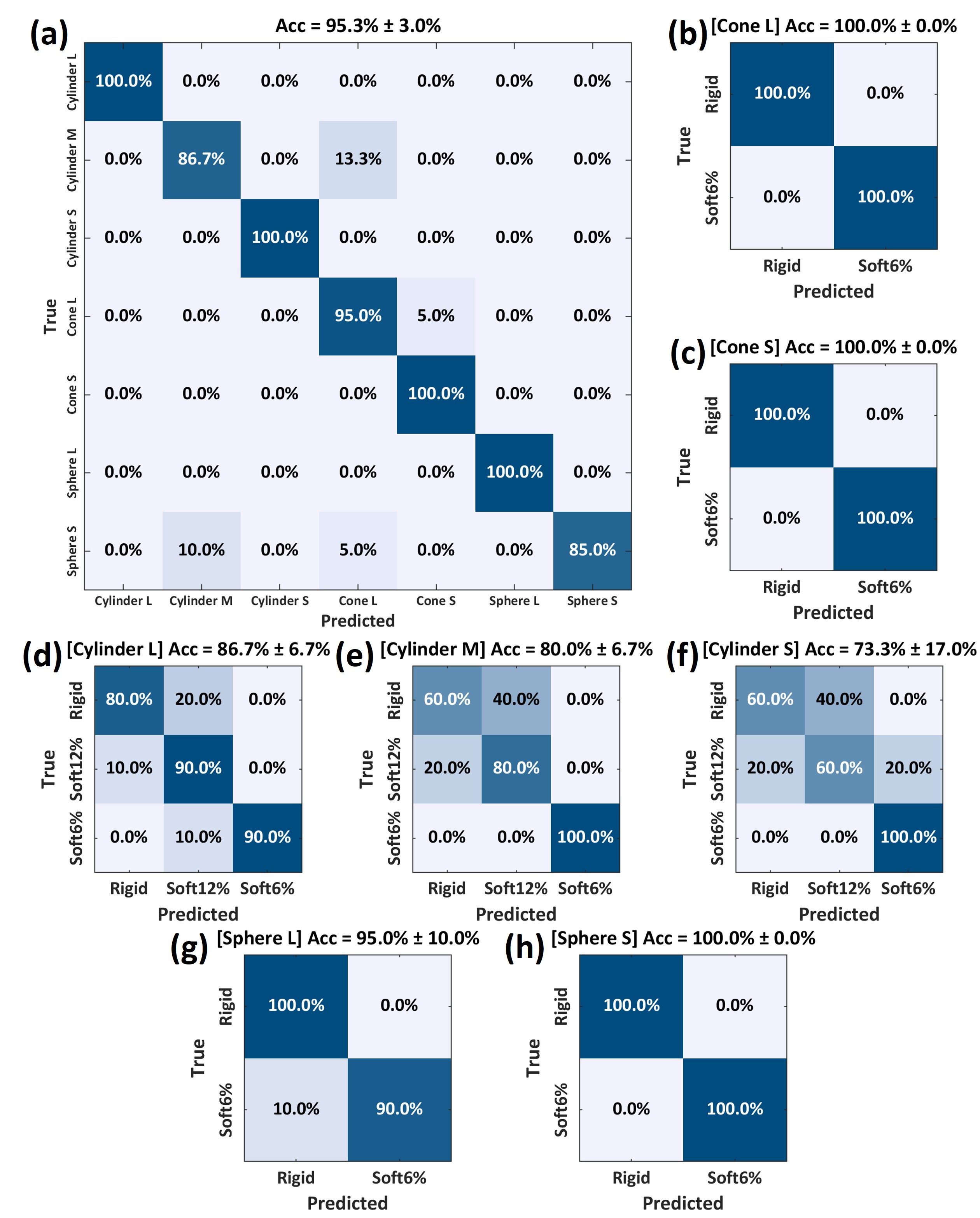}
\caption{Confusion matrices for object recognition k-NN model.
(a) Geometry.
(b)--(h) Stiffness in each geometry category.}
\label{fig_15}
\end{figure}

\section{Discussion \& Conclusion}
This study presented an integrated system for tendon-based proprioception in an anthropomorphic underactuated hand. 
The MPC-SEA was developed as a compact sensing–actuation module and integrated with the sensorless fingers and on-board electronics to form the complete system.
By utilizing an energy-based finger model, the tension profile serves as an informative grasp state indicator, enabling comprehensive interpretation of hand--object interaction throughout the entire grasp sequence.
The proposed grasp state estimation framework is also scalable to general tendon-driven underactuated finger designs with quantitative grasp state information, as long as the joint stiffness is well defined and the system energy can be modeled.
To summarize, the main contributions are as follows:
1) A generalizable grasp state estimation framework for underactuated hands that infers multiple grasp states from only a single proprioceptive tension signal, using energy-based finger modeling and an efficient rule-based algorithm.
2) Development of the MPC-SEA module distinguished by its high sensing accuracy and reliability with full integrability into compact robotic hand structures.
3) Experimental validation demonstrating the efficacy of the proposed framework and extensive demonstrations showing practical grasp functionalities.

Despite these results, several limitations remain.
First, the estimation framework can be sensitive to friction as it is not explicitly modeled. While its effect was minimized in our design by using bearings and bushings, other hardware with higher friction may experience larger estimation errors.
Second, the approach depends on appropriate design parameters; for instance, overly weak torsional springs at the joints may lead to tension variations too small for reliable proximal contact detection.
Finally, the interpretation of tension--excursion profiles is currently implemented through a rule-based algorithm, limiting robustness under unexpected or extreme grasping conditions.
Future work will explore learning-based methods to process tension profile, which may enable more generalized controllers capable of handling a wider range of situations.

\bibliographystyle{IEEEtran}
\bibliography{references}
\vspace{-4.0em}
\begin{IEEEbiography}[{\includegraphics[width=1in,height=1.25in,clip,keepaspectratio]{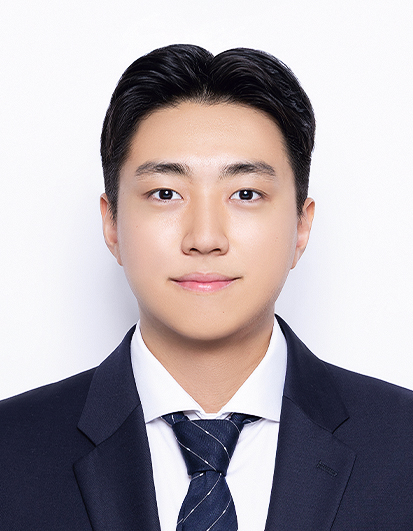}}]{Jae-Hyun Lee}
(Student Member, IEEE) received the B.S. degree in mechanical engineering and in industrial engineering (Dual degree) from Seoul National University, Seoul, Republic of Korea, in 2024. He is currently working toward the M.S. degree in mechanical engineering with the BioRobotics Laboratory, Seoul National University. His research interests include manipulation, robotic hands, and novel hardware designs.
\end{IEEEbiography}
\vspace{-4.0em}
\begin{IEEEbiography}
[{\includegraphics[width=1in,height=1.25in,clip,keepaspectratio]{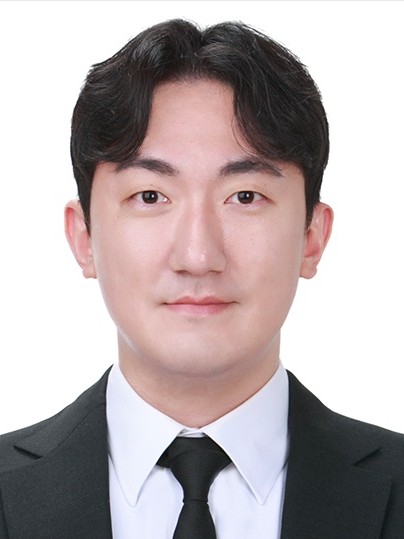}}]{Jonghoo Park}
received the B.S degree in mechanical engineering from Georgia Institute of Technology, Atlanta, GA, USA, in 2016, and M.S. and Ph.D. degrees in mechanical engineering from Seoul National University, Seoul, Republic of Korea, in 2019 and 2025, respectively. He is currently a Principal Research Engineer in Robotics Development Team at Hyundai Mobis, Seoul, Republic of Korea. His research interests include bio-inspired joints, 3D/4D printable mechanism design, and robotic hands.
\end{IEEEbiography}
\vspace{-4.0em}
\begin{IEEEbiography}
[{\includegraphics[width=1in,height=1.25in,clip,keepaspectratio]{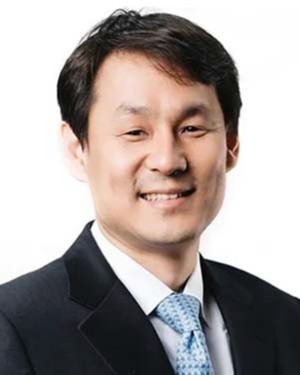}}]{Kyu-Jin Cho}
(Member, IEEE) received the B.S and M.S. degrees in mechanical engineering from Seoul National University, Seoul, Republic of Korea, in 1998 and 2000, respectively, and the Ph.D. degree in mechanical engineering from the Massachusetts Institute of Technology, Cambridge, MA, USA, in 2007. He was a Postdoctoral Fellow with Harvard Microrobotics Laboratory until 2008. He is currently a Professor of mechanical engineering and the Director of BioRobotics Laboratory, Seoul National University, and the Director of Soft Robotic Research Center. His research interests include biologically inspired robotics, soft robotics, soft wearable devices, novel mechanisms using smart structures, and rehabilitation/assistive robotics. 
\end{IEEEbiography}

\end{document}